\documentclass[3p,nopreprintline,times]{elsarticle}
\usepackage{caption}
\usepackage{subcaption}
\usepackage{color}
\usepackage{amssymb}
\usepackage{amsmath}
\usepackage{multirow}
\usepackage{lineno}

\usepackage[figuresright]{rotating}
\usepackage[pdftex,pdfpagelabels,bookmarks,hyperindex,hyperfigures]{hyperref}

\usepackage{fancyhdr}
\fancypagestyle{preprint}{\fancyhf{}\fancyfoot[L]{\it
    Preprint --
to appear in Computers and Electronics in Agriculture}}

\renewcommand{\vec}[1]{\boldsymbol{#1}} 

\begin{document}

\begin{frontmatter}


\title{Automatic moth detection from trap images for pest management}


\author{Weiguang Ding}
\author{Graham Taylor}

\address{School of Engineering\\
University of Guelph\\
50 Stone Rd.~E., Guelph, Ontario, Canada N1G 2W1\\
\{wding,gwtaylor\}@uoguelph.ca}

\begin{abstract}
Monitoring the number of insect pests is a crucial component in pheromone-based
pest management systems. In this paper,
we propose an automatic detection pipeline based on deep learning
for identifying and counting pests in images taken inside field traps.
Applied to a commercial codling moth dataset,
our method shows promising performance both qualitatively and
quantitatively. Compared to previous attempts at pest detection, our
approach uses no pest-specific engineering which enables it to adapt
to other species and environments with minimal human effort. It is
amenable to implementation on parallel hardware and therefore capable
of deployment in settings where real-time performance is required.
\end{abstract}


\end{frontmatter}



\thispagestyle{preprint}

\section{Introduction}
\label{sec:intro}

Monitoring is a crucial component in pheromone-based pest control
\cite{carde1995control, witzgall2010sex} systems.  In widely used
trap-based pest monitoring, captured digital images are analysed by
human experts for recognizing and counting pests.  Manual counting is
labour intensive, slow, expensive, and sometimes error-prone, which
precludes reaching real-time performance and cost targets.  Our goal
is to apply state-of-the-art deep learning techniques to pest
detection and counting, effectively removing the human from the loop
to achieve a completely automated, real-time pest monitoring system.


Plenty of previous work has considered insect classification.  The
past literature can be grouped along several dimensions, including
image acquisition settings, features, and classification algorithms.
In terms of image sources, many previous methods have considered
insect specimens \cite{kang2012identification, kang2014identification,
  arbuckle2001biodiversity, weeks1999automating, tofilski2004drawwing,
  wang2012new}.  Specimens are usually well preserved and imaged in an
ideal lab environment.  Thus specimen images are consistent and
captured at high resolution.  In a less ideal but more practical
scenario, some other works attempt to classify insects collected in
the wild, but imaged under laboratory conditions
\cite{larios2008automated, martinez2009dictionary,larios2010haar,
  lytle2010automated,
  al2010identification,cho2007automatic,mayo2007automatic}.  In this
case, image quality is usually worse than the specimen case, but
researchers still typically have a chance to adjust settings to
control image quality, such as imaging all of the insects under a
standard orientation or lighting.

From an algorithmic perspective,
various types of features have been used for insect classification,
including wing structures \cite{kang2012identification, kang2014identification,
arbuckle2001biodiversity, weeks1999automating, tofilski2004drawwing},
colour histogram features \cite{le2010auto, kaya2014application},
morphometric measurements \cite{fedor2008thrips, yaakob2012insect,
tofilski2004drawwing, wang2012new},
local image features
\cite{le2010auto, kaya2014application, wen2009local, wen2012image,
lu2012insect, larios2011stacked}, and
global image features \cite{xiao2009knn}.
Different classifiers were also used on top of these various feature
extraction methods,
including support vector machines (SVM)
\cite{wen2009local, wang2012new, larios2010haar},
artificial neural networks (ANN)
\cite{wang2012new, kaya2014application, fedor2008thrips},
k-nearest neighbours (KNN) \cite{xiao2009knn, wen2012image},
and ensemble methods
\cite{larios2008automated, martinez2009dictionary, wen2012image}.
In general, however, these proposed methods were not tested under
real application scenarios, for example, images from real traps deployed
for pest monitoring.

Object detection involves also localizing objects in addition to classification.
A few attempts have been made with respect to insect detection.
One option is to perform a ``sliding window'' approach,
where a classifier scans over patches at different locations of the image.
This technique was applied for inspection of bulk wheat samples
\cite{zayas1998detection}, where local patches from the original image
were represented by engineered features and classified
by discriminant analysis.
Another work on bulk grain inspection \cite{ridgway2002rapid} employed
different customized rule-based algorithms to detect different
objects, respectively. The other way of performing detection is
to first propose initial detection candidates
by performing image segmentations. These candidates are then represented by
engineered features and classified \cite{qing2012insect, yao2013segmentation}.
All of these insect detection methods are heavily engineered and work only on
specific species under specific environments,
and are not likely to be directly effective in the pest monitoring setting.


There are two main challenges in detecting pests from trap images.
The first challenge is low image quality, due to constraints such as
the cost of the imaging sensor, power consumption, and the speed by which
images can be transmitted.  This makes most of the previous work
impractical, that is, those based on high image quality and fine
structures.  The second challenge comes from inconsistencies which
are driven by many factors, including illumination, movement of the trap,
movement of the moth, camera out of focus, appearance of other objects
(such as leaves), decay or damage to the insect,
appearance of non-pest (benign) insects, etc.  These make it very hard to
design rule-based systems.  Therefore, an ideal detection method
should be capable and flexible enough to adapt to different varying
factors with a minimal amount of additional manual effort other than
manually labelled data from a daily pest monitoring program.


Apart from the insect classification/detection community, general
visual object category recognition and detection has been a mainstay
of computer vision for a long time.  Various methods and datasets
\cite{zhang2013object,andreopoulos201350} have been proposed in the
last several decades to push this area forward.  Recently,
convolutional neural networks (ConvNets)
\cite{lecun1998gradient,krizhevsky2012imagenet} and their variants
have emerged as the most effective method for object recognition and
detection, by achieving state-of-the-art performance on many well
recognized datasets \cite{ciresan2012multi, lee2014deeply,
  swersky2013multi}, and winning different object recognition
challenges \cite{russakovsky2014imagenet, krizhevsky2012imagenet,
  szegedy2014going}.

Inspired by this line of research, we adopt the popular sliding window
detection pipeline with convolutional neural networks as the image
classifier.  First, raw images are preprocessed with colour correction.
Then, trained ConvNets are applied to densely sampled image patches to
predict each patch's likelihood of containing pests.  Patches are then
filtered by non-maximum suppression, after which only those with
probabilities higher than their neighbours are preserved.  Finally,
the remaining patches are thresholded.  Patches whose probability meet
the threshold are considered as proposed detections.

This paper makes two main contributions.  First, we develop a
ConvNet-based pest detection method, that is accurate, fast, easily
extendable to other pest species, and requires minimal pre-processing
of data.
Second, we propose an evaluation metric for pest
detection borrowing ideas from the pedestrian detection literature.

\section{Data collection}
\label{sec:materials}

In this section, we describe the collection, curation, and
preprocessing of images. Details of detection performed on processed
images are provided in Section \ref{sec:det_pipeline}.

\subsection{Data acquisition}

\begin{figure}[t]
\centering
\begin{subfigure}[b]{0.5\textwidth}
\includegraphics[width=1\textwidth]{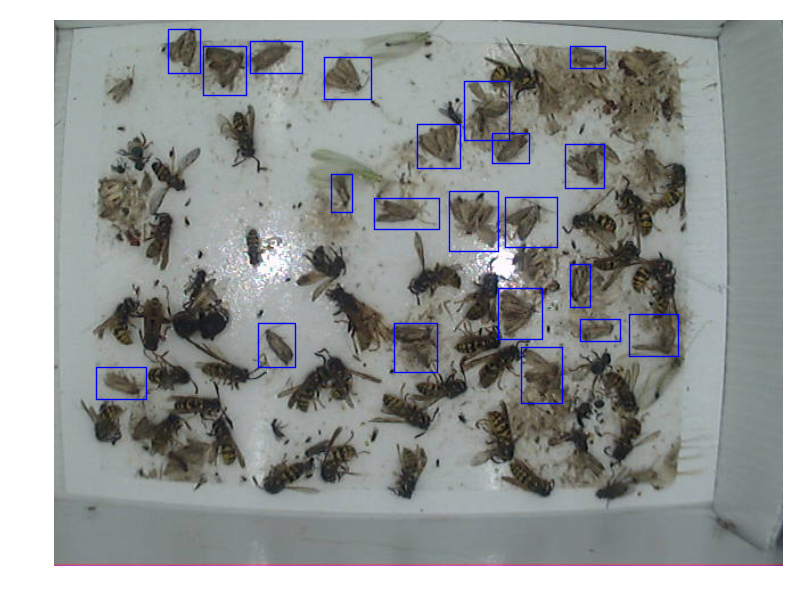}
\captionsetup{margin=10pt}
\caption{Trap containing moths. Blue rectangles indicate codling moths
  labelled by technicians.}
\label{fig:pos_img}
\end{subfigure}%
\begin{subfigure}[b]{0.5\textwidth}
\includegraphics[width=1\textwidth]{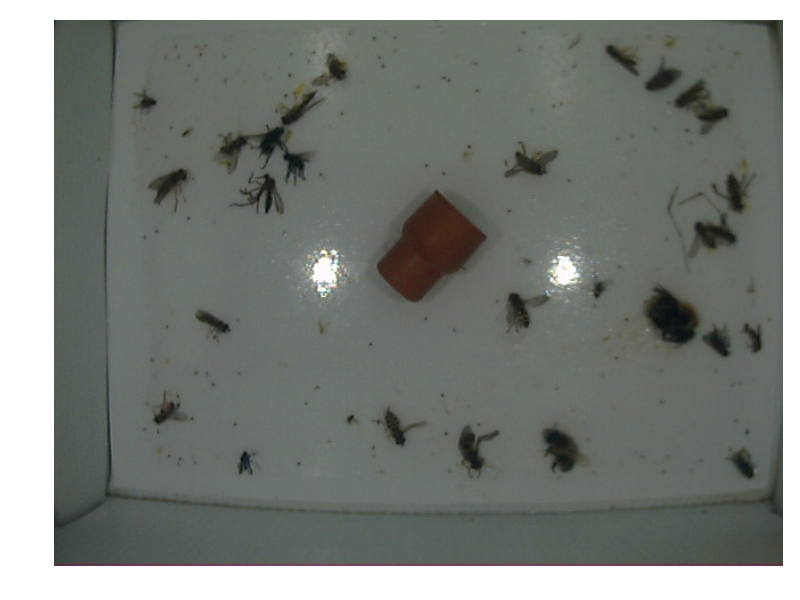}
\captionsetup{margin=10pt}
\caption{Trap containing no moths. The
  large object in the centre is a pheromone lure.}
\label{fig:neg_img}
\end{subfigure}%

\caption{Examples of images captured within the pheromone traps. Best viewed in colour.}
\end{figure}

RGB colour images are captured by pheromone traps installed at multiple
locations by a commercial provider of pheromone-based pest control
solutions, whose name is withheld by request.  The trap contains a
pheromone lure, an adhesive liner, a digital camera and a radio
transmitter.  The pheromone attracts the pest of interest into the
trap where they become stuck to the adhesive surface.  The digital
images are stored in JPEG format at 640$\times$480 resolution, and
transmitted to a remote server at fixed time point daily. Codling
moths are identified and labelled with bounding boxes
by technicians trained in entomology.
Only one image from each temporal sequence is labelled and used in this study,
so labelled images do not have temporal correlation with each other.
As a result, all of the labelled moths are unique.
Figure \ref{fig:pos_img} shows
a trap image with all the codling moth labelled with blue bounding
boxes. Figure \ref{fig:neg_img} shows an image containing no moths but
cluttered with other types of insects.
High resolution individual image patches are shown later in Figure
\ref{fig:variability}, with their characteristics analysed in Section
\ref{sec:patch_variability}.

\subsection{Dataset construction}

The set of collected images is split randomly into 3 sets: the training set,
the validation set and the test set.
After splitting, the statistics of each set is roughly the same as
the entire dataset, including the ratio between the number of images with or
without moths, and number of moths per image.
Table \ref{tab:data} provides specific statistics on
the entire dataset and the three splits subsequently constructed.

\begin{table}[t]
\centering
\begin{tabular}{cccccc}

\hline
Dataset & Total \# & \# images with moth & \# images without moth & \# moths & avg.~\# moths per image \\
\hline
Total   & 177 & 133 & 44 & 4447 & 25.1\\
Training   & 110 & 83 & 27 & 2724 & 24.8\\
Validation  & 27 & 20 & 7 & 690 & 25.6\\
Test  & 40 & 30 & 10 & 1033 &  25.8 \\
\hline
\end{tabular}
\caption{Statistics of constructed datasets.}
\label{tab:data}
\end{table}

\subsection{Preprocessing}
\label{sec:preprocessing}

\begin{figure}[t]
\centering
\begin{subfigure}[b]{0.5\textwidth}
\includegraphics[width=0.9\textwidth]{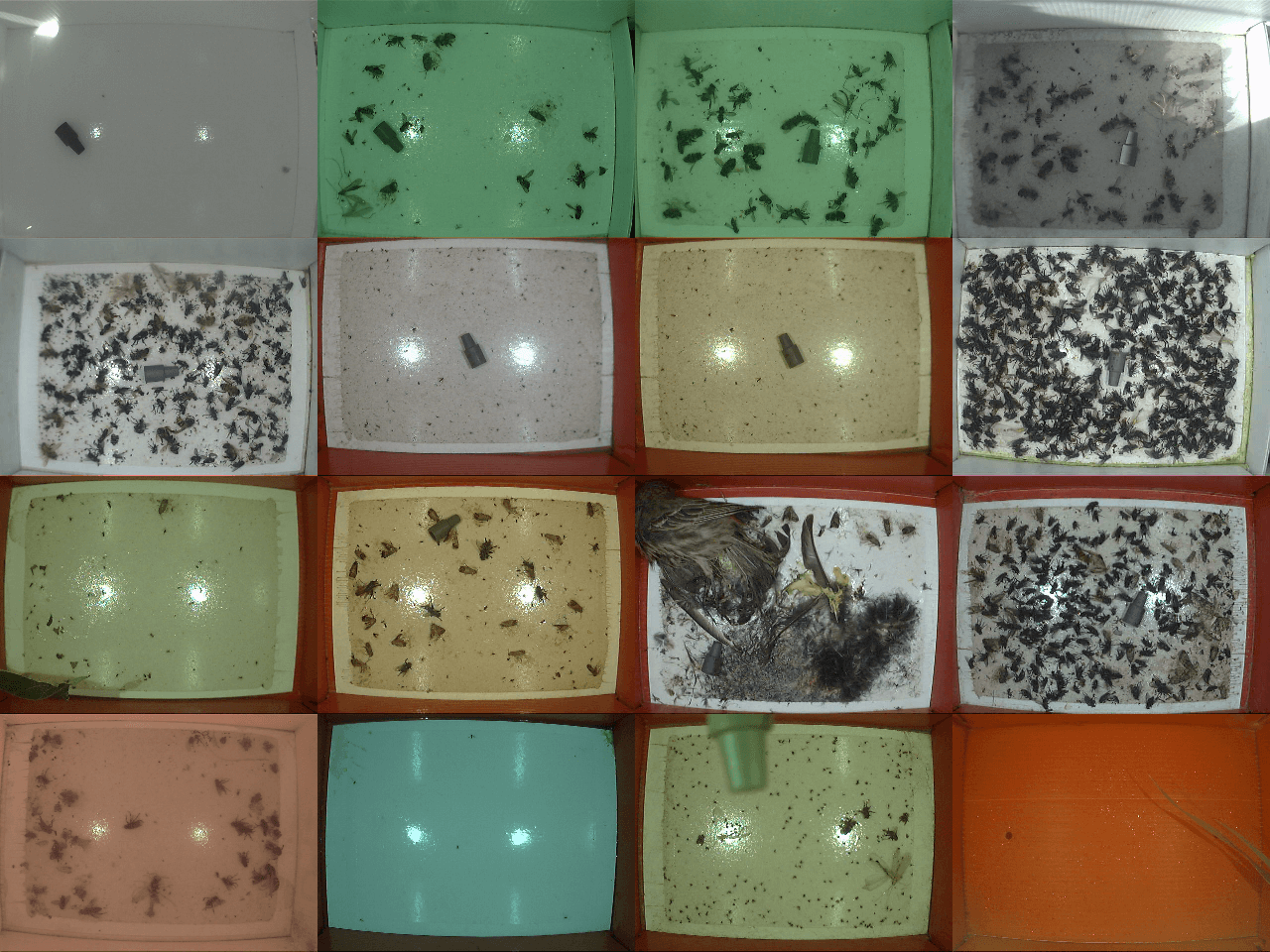}
\caption{Original images}
\label{fig:color-orig}
\end{subfigure}%
\begin{subfigure}[b]{0.5\textwidth}
\includegraphics[width=0.9\textwidth]{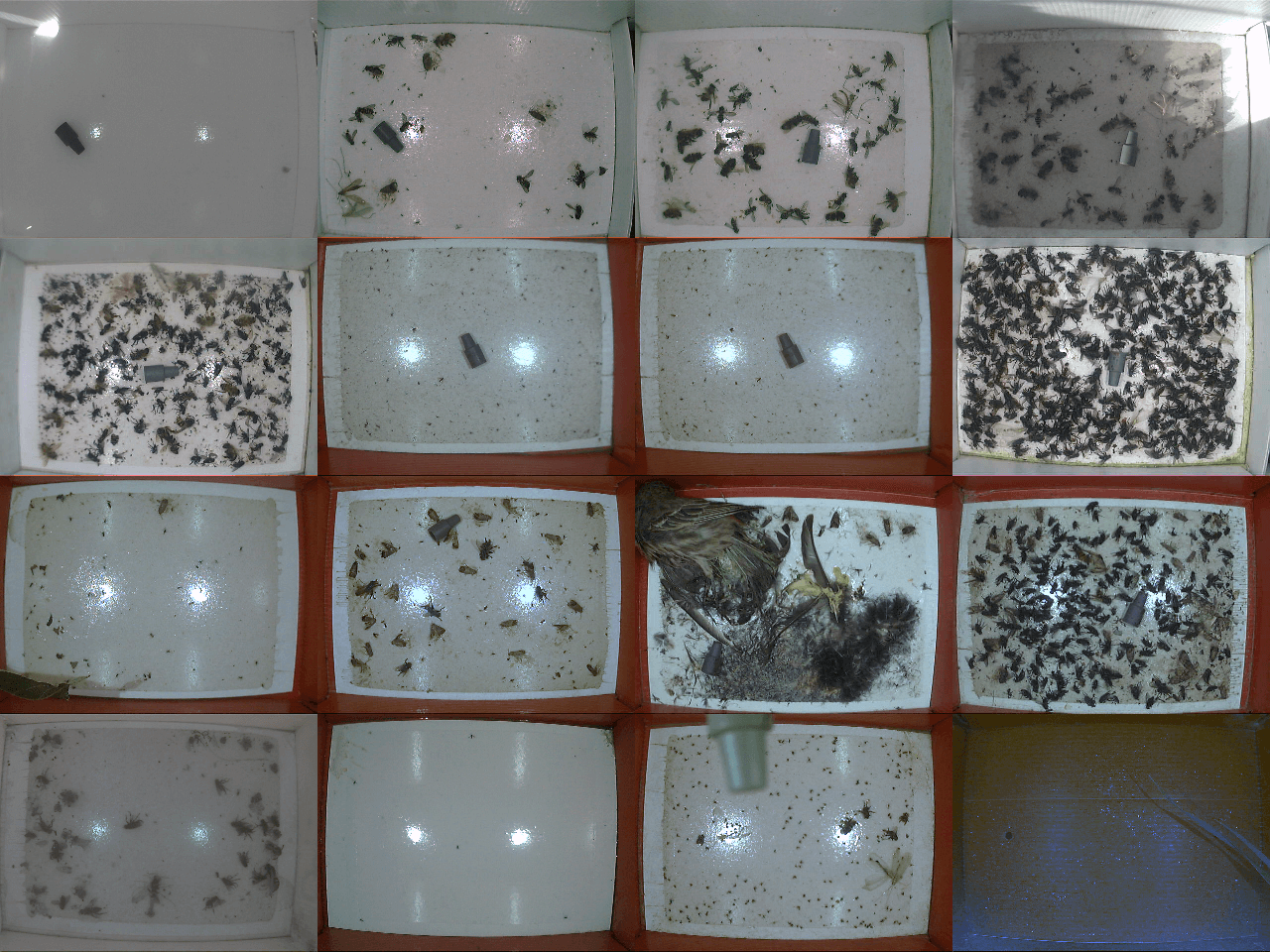}
\caption{Images processed by the ``grey world'' algorithm \cite{nikitenko2008applicability}.}
\label{fig:color-proc}
\end{subfigure}
\caption{Illustration of grey world colour correction. Best viewed in colour.}
\label{fig:color}
\end{figure}

Trap images were collected in real production environments,
which leads to different imaging conditions at different points in time.
This is most apparent in illumination, which can
be seen in Figure \ref{fig:color-orig}.
To eliminate the potential negative effects of illumination variability on detection performance,
we perform colour correction using one variant \cite{nikitenko2008applicability}
of the ``grey-world'' method. This algorithm assumes that
the average value of red (R), green (G) and blue (B) channels
should equal to each other.
Specifically, for each image, we set the gain of the R and B channels as follows:

\begin{equation}
G_{red} = \mu_{red} / \mu_{green}, ~ G_{blue} = \mu_{blue} / \mu_{green}
\end{equation}

where $\mu_{red}$, $\mu_{green}$ and $\mu_{blue}$ are the original
average intensities of the red, green and blue channels, respectively.
$G_{red}$ and $G_{blue}$ are multiplicative gains applied to the pixel intensity values
of the red and blue channels, respectively.
Figure \ref{fig:color-proc} shows images processed by the grey-world algorithm.
We see that the images are white-balanced to have similar illumination, but still
maintain rich colour information which can be a useful cue for
detection downstream.
In this paper, all images are white-balanced prior to detection.

\section{Detection pipeline}
\label{sec:det_pipeline}
\begin{figure}[t]
\centering
\includegraphics[width=0.8\textwidth]{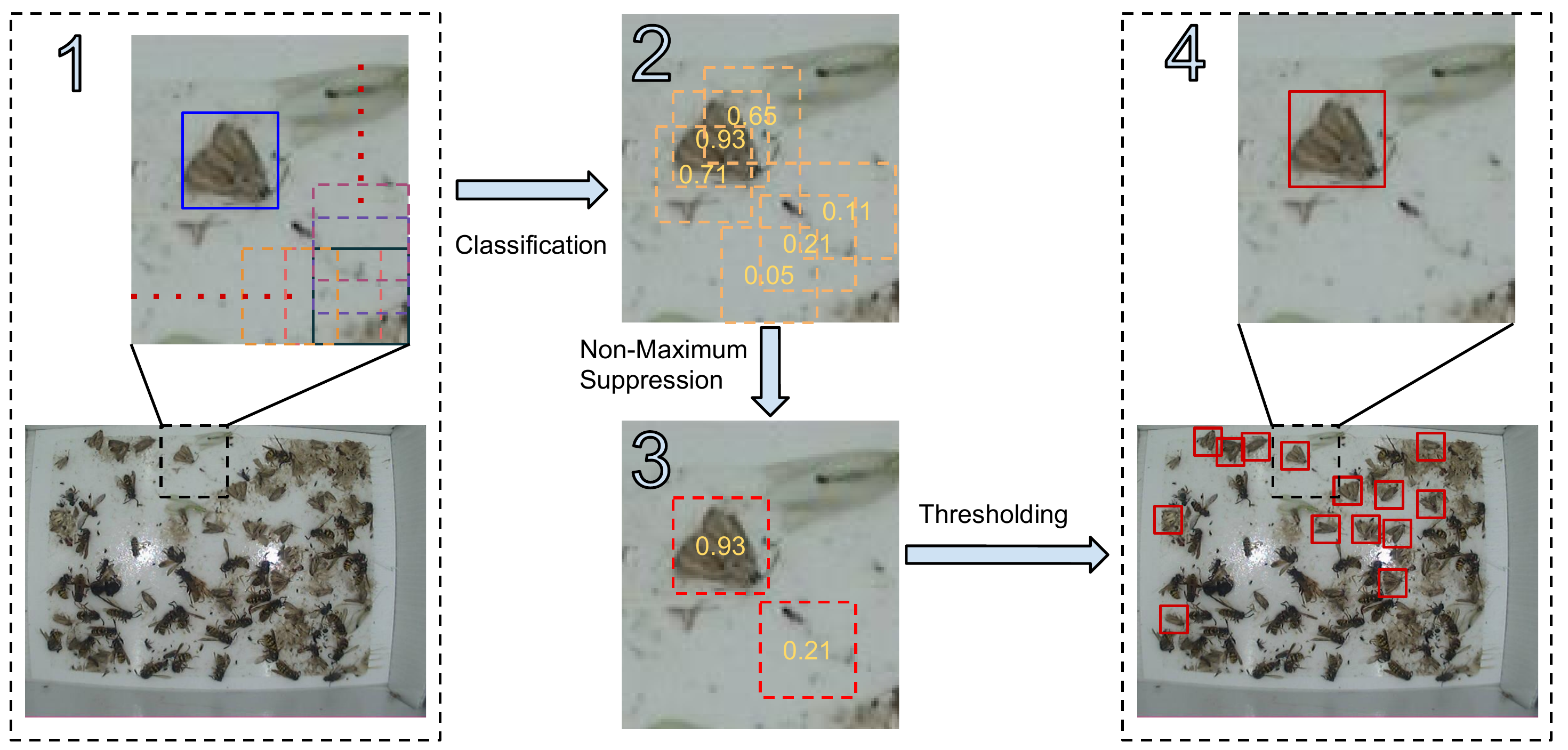}
\caption{Illustration of the detection pipeline. Best viewed in colour.}
\label{fig:pipeline}
\end{figure}

The automatic detection pipeline involves several steps, as shown in
Figure \ref{fig:pipeline}.  We take a sliding window approach, where a
trained image classifier is applied to local windows at different
locations of the entire image. The classifier's output is a single
scalar $p \in [0, 1]$, which represents the probability that a
particular patch contains a codling moth.  These patches are regularly
and densely arranged over the image, and thus largely overlapping.
Therefore, we perform non-maximum suppression (NMS) to retain only the
windows whose respective probability is locally maximal. The remaining
boxes are then thresholded, such that only patches over a certain
probability are kept. The location of these patches with their
respective probabilities (confidence scores) are the final outputs of
the detection pipeline. We now discuss each of these stages in more
detail.

\subsection{Convolutional neural networks}

\begin{figure}[t]
\centering
\includegraphics[width=0.9\textwidth]{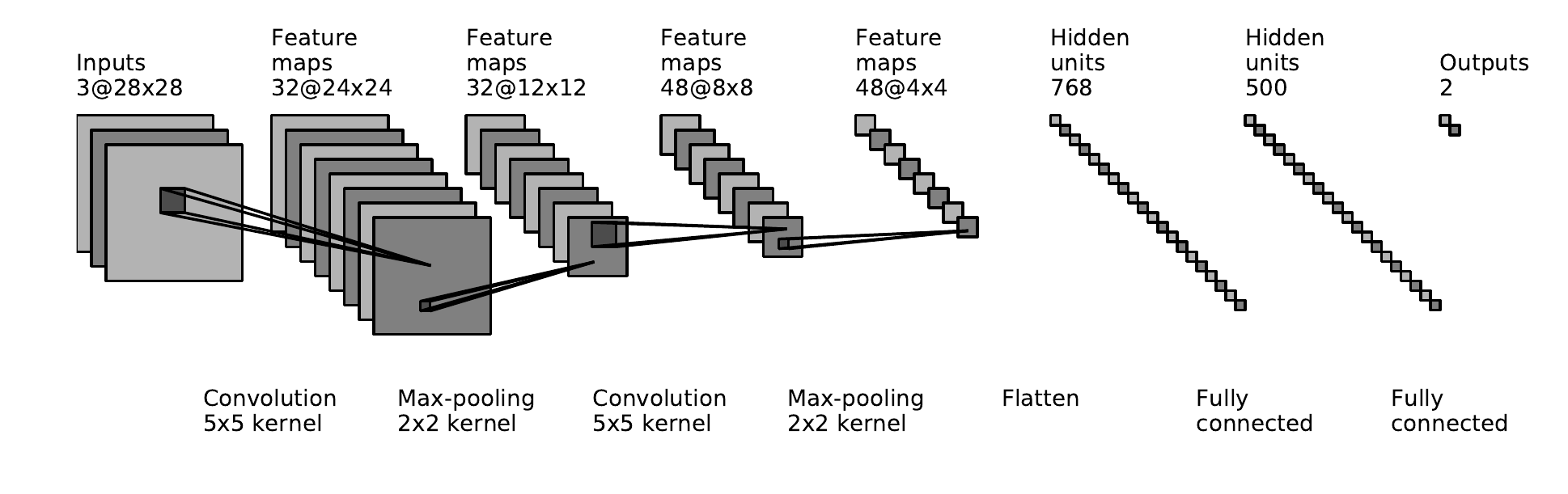}
\caption{The convolutional neural network architecture used in our experiments.}
\label{fig:convnet}
\end{figure}

In a sliding window approach, the detection problem breaks down into classifying each local patch, which is performed by the image classifier $C$, a mapping from the image $I$ to a probability $p$: $C: I \mapsto p$.
We adopt convolutional neural network \cite{lecun1998gradient}
(ConvNet) as our image classifier, as it is the most popular and best performing classifier for image recognition in both large scale \cite{krizhevsky2012imagenet, szegedy2014going} and small scale \cite{lin2013network,krizhevsky2009learning} problems. It is also very fast to deploy, and amenable to parallel hardware.
Specifically, we used a network structure similar to Lenet5 \cite{lecun1998gradient}. As shown in Figure \ref{fig:convnet}, our network contains 2 convolutional layers, 2 max-pooling layers, and 2 fully connected layers (described below).
Before applying the ConvNet, each dimension of the input patch\footnote{Note that we distinguish patch-based normalization, as described here, with whole image normalization, described in Section \ref{sec:preprocessing}.} is normalized to have zero mean and unit variance.

\subsubsection{Convolutional layers}
\label{conv_layer}

A convolutional layer applies a linear filterbank and element-wise nonlinearity to its input ``feature maps'', transforming them to a different set of feature maps. By applying convolutional layers several times, we can extract increasingly high-level feature representations of the input, at the same time preserving their spatial relationship. At the first layer, the input feature maps are simply the channels of the input. At subsequent layers, these represent more abstract transformations of the image.
A convolutional layer is a special case of a fully connected layer, introduced in Subsection \ref{sec:fc_layer}, where only local connections are have non-zero values, and weights are tied at all locations. Local connectivity is implemented efficiently by applying convolution:

\begin{equation}
\vec{h}^{l}_k = \varphi \left( \sum_{m} \vec{W}^{l}_{m,k} \vec{h}^{l-1}_{m} + \vec{b}^l_k \right) ~,
\end{equation}
where $l$ is the layer index; $m$ is the index of input feature maps;
$k$ is the index of output feature maps; input $\vec{h}^{l-1}_m$ is the $m$th  feature map at layer $l-1$; output $\vec{h}^{l}_k$ the $k$th feature map at layer $l$; $\vec{W}$ is the convolutional weight tensor; $\vec{b}$ is the bias term; and we choose the element-wise nonlinearity $\varphi(\cdot)$ to be the rectified linear unit (RELU) \cite{grangier2009deep} function.

\subsubsection{Max-pooling layers}
\label{sec:ss_layer}

Each convolutional layer is followed by a max-pooling layer.
This layer applies local pooling operations to its input feature maps,
by only preserving the maximum value within a local receptive field
and discarding all other values. It is similar to a convolutional layer in the
sense that both operate locally.
Applying max-pooling layers has 2 major benefits:
1) reducing the number of free parameters, and
2) introducing a small amount of translational invariance into the network.

\subsubsection{Fully connected layers}
\label{sec:fc_layer}

The last two layers in our ConvNet are fully connected. These are
the kind of layers found in standard feed-forward neural networks.
The first fully connected layer flattens (vectorizes) all of the feature maps
after the last max-pooling layer, treating this
one-dimensional vector as a feature representation of the whole image.
The second fully connected layer is parameterized like a linear classifier.
Mathematically, the fully connected layer can be written as:

\begin{equation}
\vec{h}^l = \varphi(\vec{W}^l \vec{h}^{l-1} + \vec{b}^l),
\end{equation}
where $l$ is the layer index;
input $\vec{h}^{l-1}$ is the vector representation at layer $l-1$;
$\vec{h}^{l}$ is the output of layer $l$;
$\vec{W}$ is the weight matrix; $\vec{b}$ is the bias vector;
and $\varphi(\cdot)$ is an element-wise nonlinear function:
we choose RELUs for the fully connected hidden layer and
a softmax for the output layer.

For a more detailed explanation of convolutional neural networks,
we refer the reader to
\cite{lecun1998gradient,krizhevsky2012imagenet,stanford2015convolutional}.

\subsection{Non-maximum suppression}
\label{sec:nms}

After applying the ConvNet in a sliding window fashion,
we obtain probabilities associated with each densely sampled patch.
If we simply applied thresholding at this point, we would get many overlapping detections.
This problem is commonly solved using non-maximum suppression (NMS) which aims to retain only patches with locally maximal probability. We adopted a strategy similar to \cite{felzenszwalb2010object}. Specifically, we first sort all the detections according to their probability.
Then, from high to low probability, we look at each detection and remove other bounding boxes that overlap at least 10\% with the current detection. After this greedy process, we generate final detection outputs as shown in Figure \ref{fig:pipeline} and later in Section \ref{sec:results_qualitative}.

\section{Experiments and evaluation}
\label{experiments}

We next introduce how we performed the experiments and the evaluation protocol.

\subsection{Classifier training}

The classifier is trained on generated patches of different sizes,
detailed in Section \ref{sec:patch}.
Minibatch stochastic gradient descent (SGD) with momentum
\cite{lecun2012efficient} was used to train the ConvNet.
The gradient is estimated with the well known back-propagation algorithm
\cite{rumelhart1985learning}.
We used a fixed learning rate of 0.002, a fixed minibatch size of 256,
and a fixed momentum coefficient of 0.9.
The validation set is used for monitoring the training process and selecting hyper-parameters.
We report performance using a classifier whose parameters are chosen according to
the best observed validation set accuracy.
The filters and fully-connected weight matrices of the ConvNets are initialized with values selected from a uniform random distribution on an interval that is a function of the number of pre-synaptic and post-synaptic units (see \cite{glorot2010understanding} for more detail).

\subsection{Training data extraction}
\label{sec:patch}

In the sliding window classification pipeline, the classifier takes a
local window as its input. Therefore we need to extract small local
patches from the original high-resolution to train the
classifier. This is performed in a memory-efficient manner, using
pointer arithmetic to create ``views'' to the data as opposed to
storing all patches in memory.

\begin{figure}[t]
\begin{subfigure}[b]{0.33\textwidth}
\centering
\includegraphics[width=0.95\textwidth]{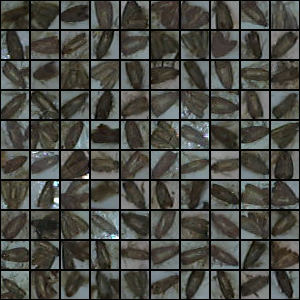}
\caption{Patches containing a moth}
\label{fig:patches_pos}
\end{subfigure}%
\begin{subfigure}[b]{0.33\textwidth}
\centering
\includegraphics[width=0.95\textwidth]{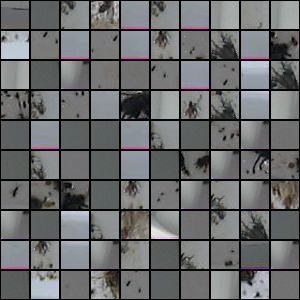}
\caption{Initial negative patches}
\label{fig:patches_neg}
\end{subfigure}%
\begin{subfigure}[b]{0.33\textwidth}
\centering
\includegraphics[width=0.95\textwidth]{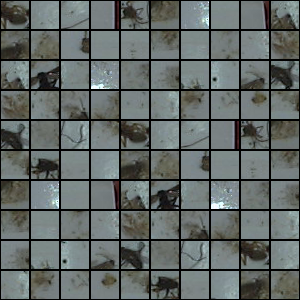}
\caption{Negative patches after bootstrapping}
\label{fig:patches_new_neg}
\end{subfigure}%
\caption{Some examples of patches used for training the classifier.}
\end{figure}

\subsubsection{Positive patches}

Here, ``positive patch'' refers to patches derived from manually labelled
bounding boxes, where each one represents a codling moth.
As the the ConvNet processes square inputs, we ignored the original aspect
ratio of the manually labelled bounding boxes, and took the square region
having the same centre as the original rectangular bounding box.
Figure \ref{fig:patches_pos} shows positive patches extracted from the training set.

\subsubsection{Negative patches}

It would be difficult to cover all the kinds of false positives that
may arise by simply sampling the regions not covered by the
labelled bounding boxes. This is because the area of regions not
containing moths is much larger than the area covered by the bounding
boxes. On images which are not very cluttered, most of the
``negative'' area is uninteresting (e.g.~trap liner).

Thus, to obtain negative training examples, we intentionally take
``hard'' patches, meaning those which contain texture.  Specifically,
we apply the Canny edge detector \cite{canny1986computational} to find
patches in ``negative images'', i.e.,~those that do not contain any
moths. We set the threshold such that the number of negative patches
roughly matches the number of labelled moths. Figure
\ref{fig:patches_neg} shows a random sample of negative patches.

\subsubsection{Bootstrapping}

After the initial set of negative patches are extracted, we use a
bootstrapping approach to find useful negative training patches that
can make the classifier more discriminative.  In the first round of
training, the initially generated patches are used to train the
classifier. At test time, the false positive patches from the training
set are collected, and we isolate the 6000 negative patches with
highest probability assigned by the classifier. These are merged with
the initially generated patches to form a new dataset for a second
stage of training.  One could potentially use more rounds of
bootstrapping to collect more informative negative patches, but we
found that including more than two training stages does not improve
performance. Figure \ref{fig:patches_new_neg} shows
randomly sampled patches collected in the test phase after one stage
of training.
For the validation set, the number of patches we collect is
proportional to the number of images in the validation set.

\subsection{Data augmentation}
\label{sec:data_aug}

\begin{figure}[t]
\centering
\includegraphics[width=0.95\textwidth]{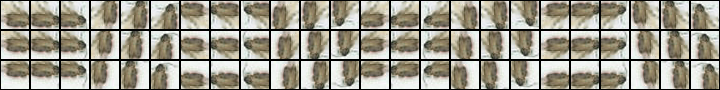}
\caption{
Data augmentation from a single example by translating, rotating, and
flipping. Note that all copies retain the same (positive) label.
}
\label{fig:augment}
\end{figure}

For machine learning-based methods, it is usually the case that the
larger the dataset, the better the generalization performance.  In our
case, the amount of training data, which is represented by the number
of training patches, is much smaller than standard small-scale image
classification datasets \cite{krizhevsky2009learning,lecun1998mnist}
frequently used by the deep learning community, which have on the
order of 50,000 training examples.  Therefore, we performed data
augmentation to increase the number of images for training, and also
incorporate invariance to basic geometric transformations into the
classifier.  Based on the ``top-view'' nature of the trap images, a
certain patch will not change its class label when it is slightly
translated, flipped, or rotated.  Therefore, we apply these simple
geometric transformations to the original patches to increase the
number of training examples.  For each patch, we create 8 translated
copies by shifting $\pm$ 3 pixels horizontally, vertically, and
diagonally. We also create versions which are flipped across the
horizontal- and vertical-axes. Finally, we create 4 rotated copies by
rotating the original by 0, 90, 180 and 270 degrees.  This produces 72
augmented patches from one original.  Figure \ref{fig:augment} shows
the augmented versions which are produced from a single example.

\subsection{Detection}
At the detection stage, we need to set the stride, which means the
distance between adjacent sliding windows.  A smaller stride means
denser patch coverage, which lead to better localization of the moths,
but also requires more computation.  As a trade-off, we set the stride
to be $\frac{1}{4}$ the size of a patch.

\subsection{Evaluation protocol}

Pest detection is still a relatively ``niche'' area of computer vision
and therefore there is no standard evaluation protocol defined. We
decided to adopt a protocol inspired by standardization within the
pedestrian detection community. A complete overview is provided in
\cite{dollar2012pedestrian} which we summarize below.

\subsubsection{Matching detections with ground truth}
We evaluate detection performance based on the statistics of
misdetections, correct detections and false positives.  Here, a
misdetection refers to a manually labelled region which is missed by
the algorithm, and a false positive refers a bounding box proposed by
the algorithm which does not correspond to any manually labelled region.
To determine if a bounding box proposed by the detector
is a correct detection or a misdetection, we determine its
correspondence to a a manually labelled bounding box by calculating the
 intersection-over-minimum (IOMin) heuristic:
\begin{equation}
A(BB_{dt}, BB_{gt}) = \frac{area(BB_{dt} \cap BB_{gt})}{min(area(BB_{dt}), area(BB_{gt}))} ~,
\end{equation}
where $BB_{dt}$ represents a bounding box proposed by the algorithm (a
detection) and $BB_{gt}$ represents a ground truth bounding box.  We
consider a specific ground truth bounding box, $BB_{gt}^i$, to be
correctly detected when there exists a detection, $BB_{dt}^j$, such
that $A(BB_{dt}^j, BB_{gt}^i) > 0.5$.  Otherwise, that ground truth
bounding box is considered to be a misdetection.  When multiple
$BB_{dt}$'s satisfy the condition, the one with the highest
probability under the classifier is chosen. After this is performed
for all of the $BB_{gt}$, the remaining unmatched $BB_{dt}$ are
considered to be false positives.

Our evaluation metric differs from that used by the pedestrian
detection in that we use IOMin in place of the more popular Jaccard
index, also called intersection-over-union (IOU). This is because of
potential shape mismatches between the ground truth and detections.
The ground truth bounding boxes are all rectangles, but our classifier
outputs probabilities over square patches.  In the case when the
ground truth rectangle is nearly square, the IOU works well.  In the
case where the ground truth rectangle is tall or wide, however, the
IOU tends to be small no matter how good the detection is.  On the
contrary, the IOMin heuristic performs well for both cases.

\subsubsection{Object level evaluation}

Based on the statistics of correct detections (also known as true positives), misdetections (also known as false negatives) and
false positives, we could evaluate the performance at two levels:
(1) object level, where the focus is on the performance
of detecting individual moths; and
(2) image level, where the focus is on determining
whether or not an image contains any moths.

At the object level, we use five threshold-dependent measures: miss rate, false positives per image (FPPI), precision, recall, and $F_{\beta}$ score:
\begin{align}
\text{miss rate} & = \frac{\text{number of misdetections}}{\text{total
                   number of moths}} \\
\text{FPPI} & = \frac{\text{number of false positives}}{\text{total number of images}} \\
\text{precision} & = \frac{\text{number of correct
                   detections}}{\text{total number of detections}} \\
\text{recall} & = \frac{\text{number of correct
                detections}}{\text{total number of moths}} \\
F_{\beta} & = (1 + \beta^2) \frac{\text{precision} \cdot
            \text{recall}}{\beta^2 \cdot \text{precision} +
            \text{recall}} \label{fbeta}
\end{align}

All quantities, with the exception of FPPI, are calculated by including
correspondences across the entire dataset, and do not represent averages
over individual images.

There are two pairs of metrics that measure the trade-off between
reducing misdetections and reducing false positives: miss rate
vs.~FPPI, and precision vs.~recall.  Miss rate vs. FPPI is a common
performance measure in the pedestrian detection community
\cite{dollar2012pedestrian}.  It gives an estimate of the system
accuracy under certain tolerances specified by the number of false
positives.  Similarly, a precision vs.~recall plot shows the trade-off
between increasing the accuracy of detection and reducing misdetections.

The $F_{\beta}$ score is simply a measure which aims to weight the
importance of precision and recall at a single operating point along
the precision-recall curve. The larger it is, the better the
performance.  The parameter $\beta$ adjusts the importance between
precision and recall. In this paper, we consider detecting all moths
more important than reducing false positives\footnote{In this domain,
  the cost of not responding to a potential pest problem outweighs
  that of unnecessarily applying a treatment.}, and therefore we more
heavily weight recall, setting $\beta=2$ for all reported results.

Of course, there will be one $F_{\beta}$ score for each operating
point (threshold) of the system. To summarize the information conveyed
by the miss rate vs.~FPPI and precision vs.~recall plots by a single
value, we employ two scalar performance measures: (1) log-average miss
rate when FPPI is in the range [1, 10], and (2) area under the
precision-recall curve (AUC).

\subsubsection{Image level evaluation}
Image level performance evaluation is considered for the scenario of
semi-automatic detection, where the algorithm proposes images for a
technician to inspect and safely ignores images that do not contain
any moths. In this setting, the algorithm simply needs to make a proposal of
``moths'' or ``no moths'' per image regardless of \emph{how many}
moths it believes are present. Here we will call a ``true moth'' image
an image that contains at least one moth, and a ``no moth'' image an
image that contains no moths. Similar to Object Level Evaluation,
there are five threshold-dependent measures: sensitivity (synonymous
with recall), specificity, precision, and $F_{\beta}$ score:

\begin{align}
\text{sensitivity} = \text{recall} &= \frac{\text{number of correctly
                    identified true moth images}}{\text{total
                     number of true moth images}}\\
\text{specificity} &= \frac{\text{number of correctly identified
                     no moth images}}{\text{total number
                     of no moth images}}\\
\text{precision} &= \frac{\text{number of correctly identified
                     true moth images}}{\text{total
                   number of moth image proposals}}
\end{align}
\noindent where $F_{\beta}$ is defined in Eq.~\ref{fbeta}. Similar to
object level evaluation, there are also two pairs of trade-offs: sensitivity vs.~specificity,
and precision vs.~recall. For scalar performance measures,
we calculate the AUC for both of these curves.

\section{Results}
\label{sec:results}

In this section, we first give some visual (qualitative) results and
then describe the results of the performance evaluation introduced
in Section \ref{experiments}.

\subsection{Qualitative}
\label{sec:results_qualitative}
\begin{figure}[t]
\centering
\begin{subfigure}[b]{\textwidth}
\centering
\includegraphics[width=0.9\textwidth]{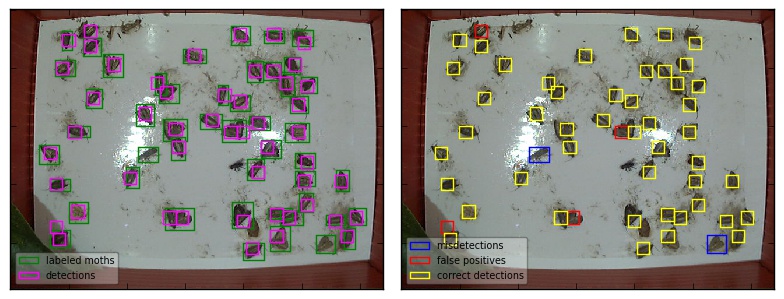}
\caption{ConvNet with input size $21\times21$.}
\label{fig:det_examples_objthresh}
\end{subfigure}%

\begin{subfigure}[b]{\textwidth}
\centering
\includegraphics[width=0.9\textwidth]{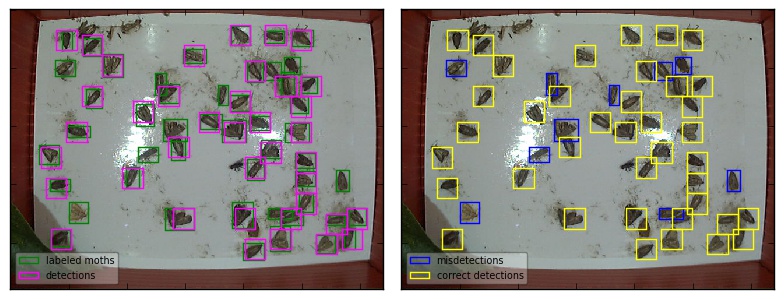}
\caption{ConvNet with input size $35\times35$.}
\label{fig:det_examples_imgthresh}
\end{subfigure}%
\caption{Visual example of detection results.
Panel (a) shows the best-performing
classifier at the object level.
Panel (b) shows the best-performing
classifier at the image level. Best viewed in colour.}
\label{fig:det_examples}
\end{figure}

\begin{figure}[t]
\centering
\includegraphics[width=0.45\textwidth]{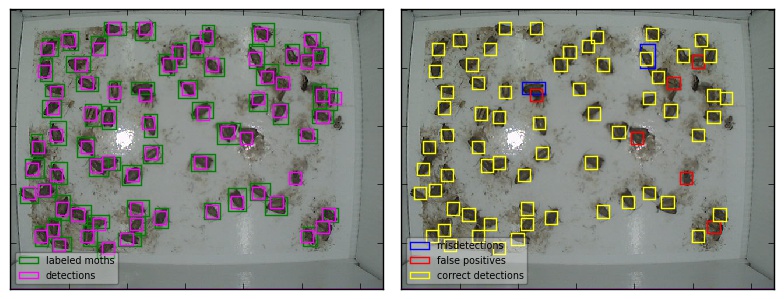}
~
\includegraphics[width=0.45\textwidth]{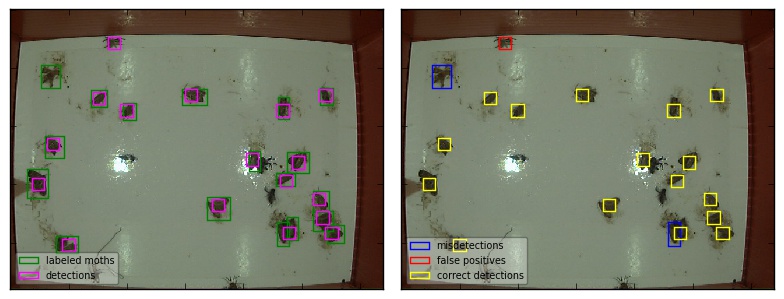}

\includegraphics[width=0.45\textwidth]{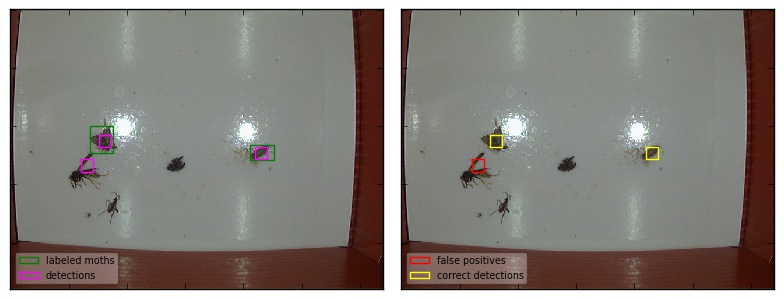}
~
\includegraphics[width=0.45\textwidth]{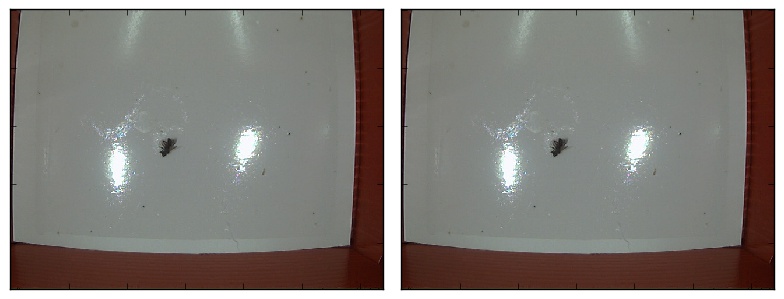}

\includegraphics[width=0.45\textwidth]{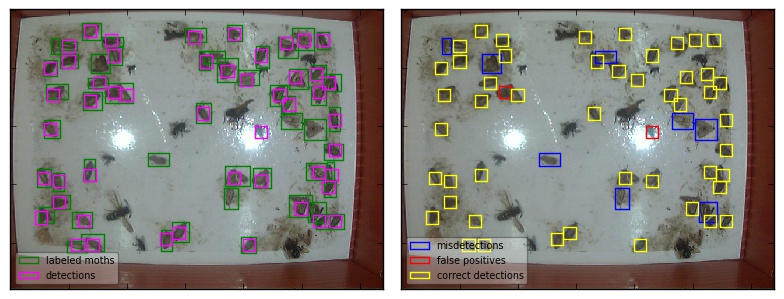}
~
\includegraphics[width=0.45\textwidth]{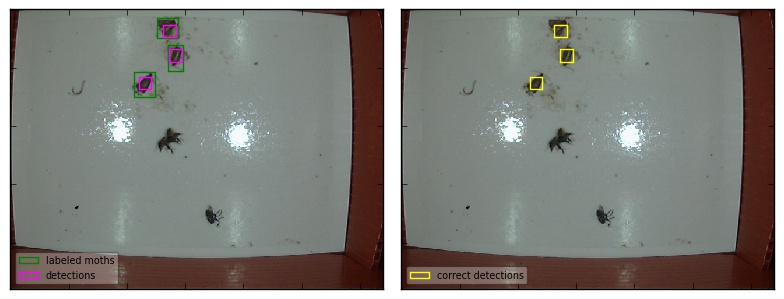}

\includegraphics[width=0.45\textwidth]{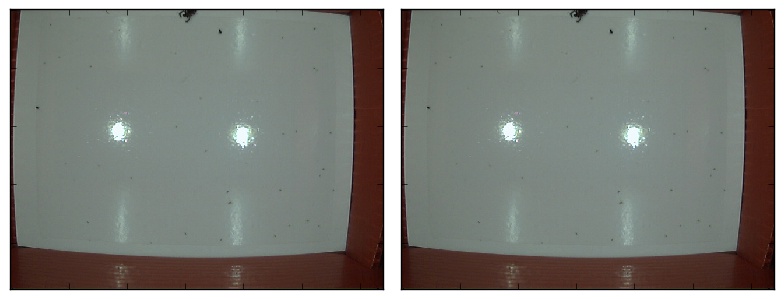}
~
\includegraphics[width=0.45\textwidth]{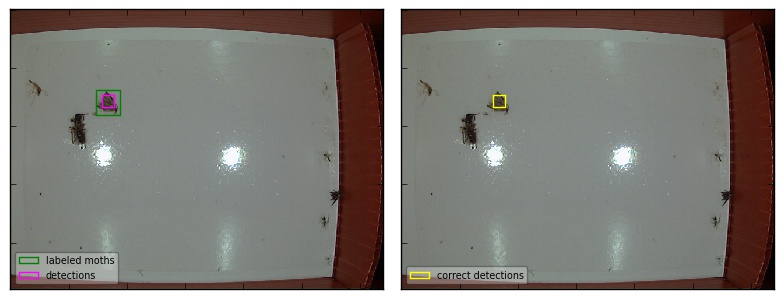}

\includegraphics[width=0.45\textwidth]{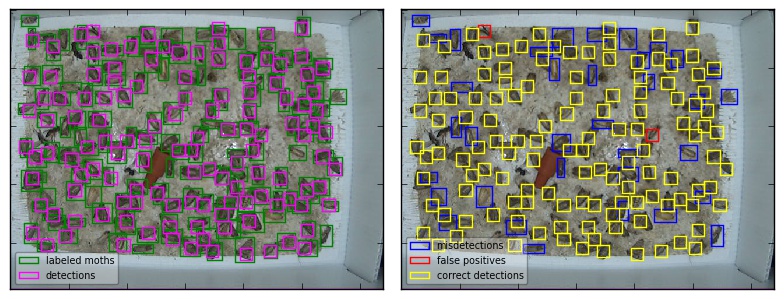}
~
\includegraphics[width=0.45\textwidth]{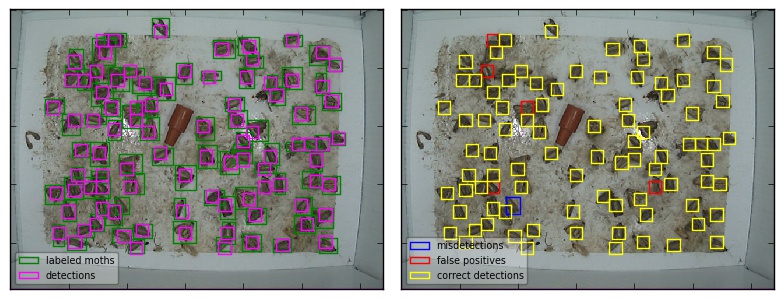}

\includegraphics[width=0.45\textwidth]{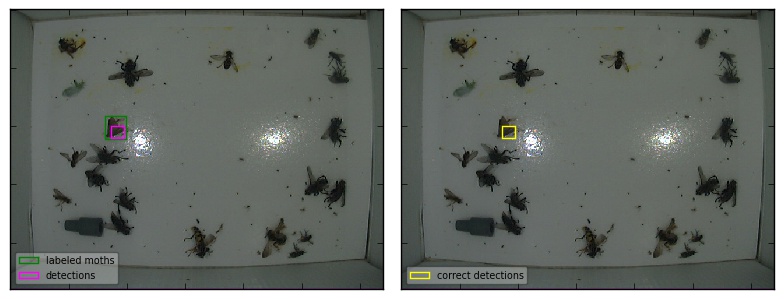}
~
\includegraphics[width=0.45\textwidth]{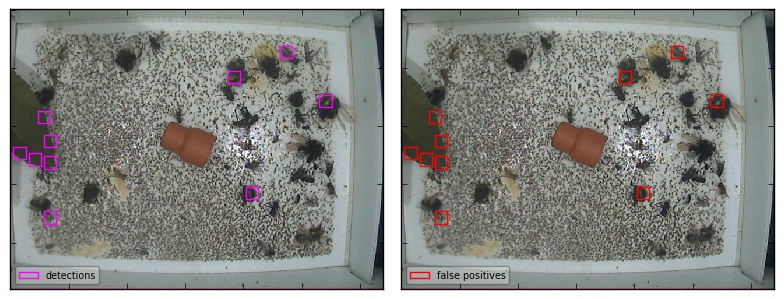}

\includegraphics[width=0.45\textwidth]{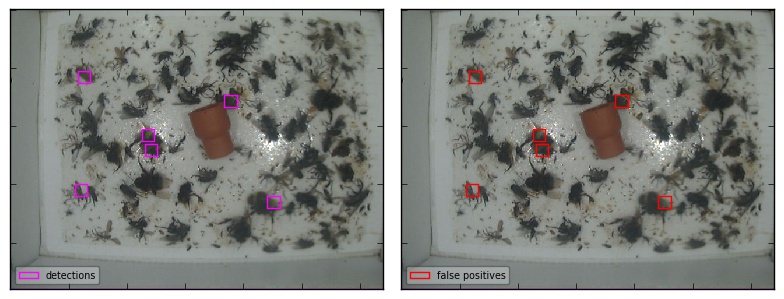}
~
\includegraphics[width=0.45\textwidth]{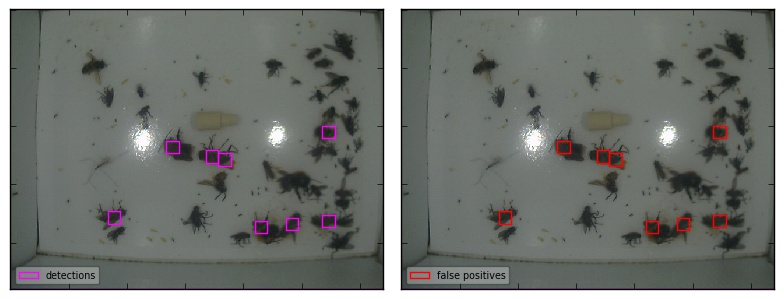}

\caption{More examples of ConvNet with input size $21\times21$. Best viewed in colour.}
\label{fig:more_det_examples_objthresh}
\end{figure}

\begin{figure}[t]
\centering
\includegraphics[width=0.45\textwidth]{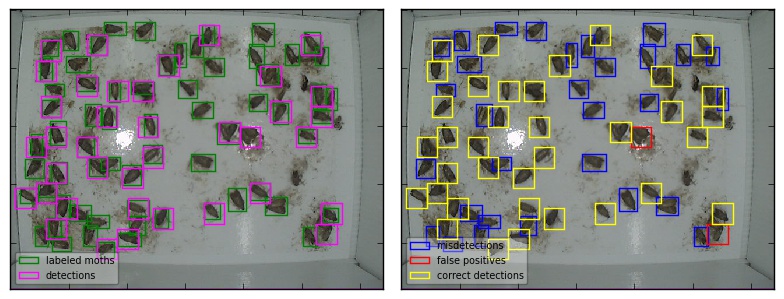}
~
\includegraphics[width=0.45\textwidth]{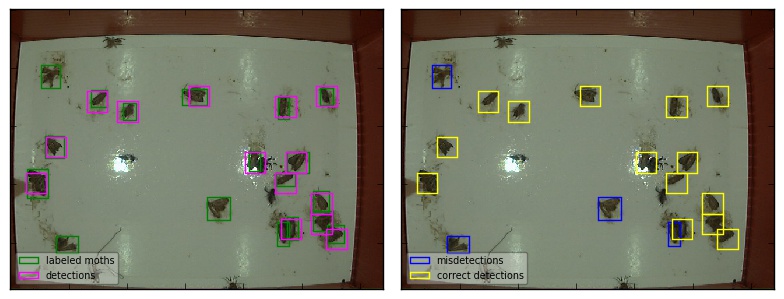}

\includegraphics[width=0.45\textwidth]{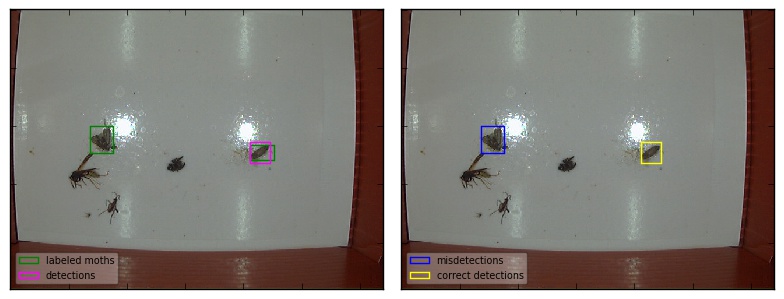}
~
\includegraphics[width=0.45\textwidth]{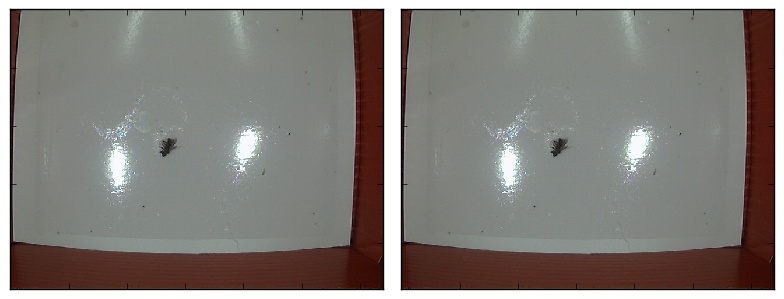}

\includegraphics[width=0.45\textwidth]{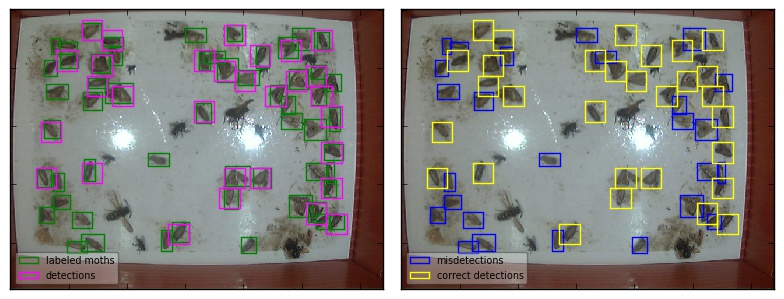}
~
\includegraphics[width=0.45\textwidth]{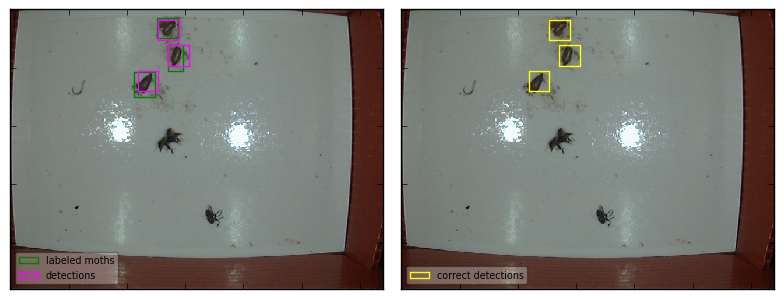}

\includegraphics[width=0.45\textwidth]{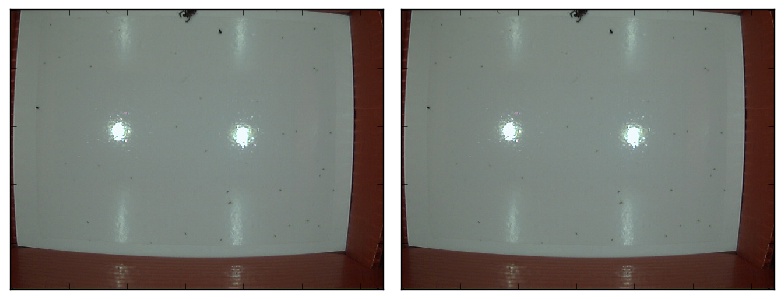}
~
\includegraphics[width=0.45\textwidth]{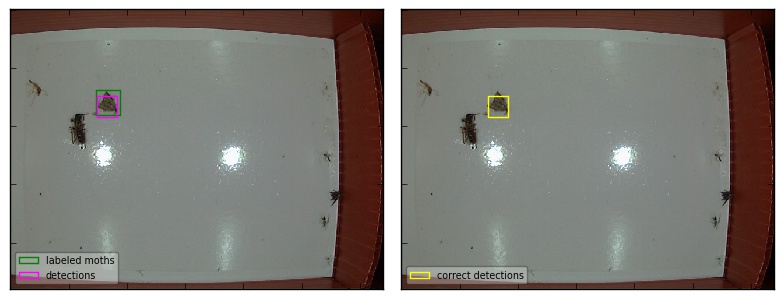}

\includegraphics[width=0.45\textwidth]{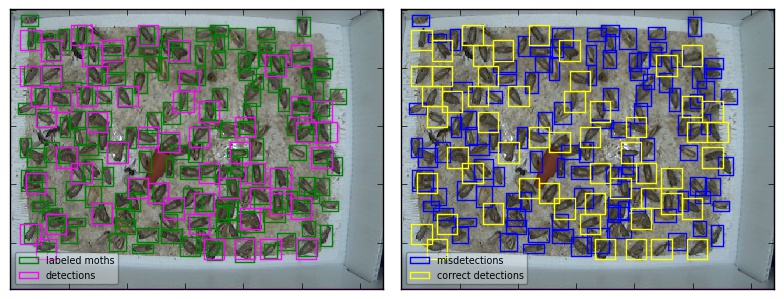}
~
\includegraphics[width=0.45\textwidth]{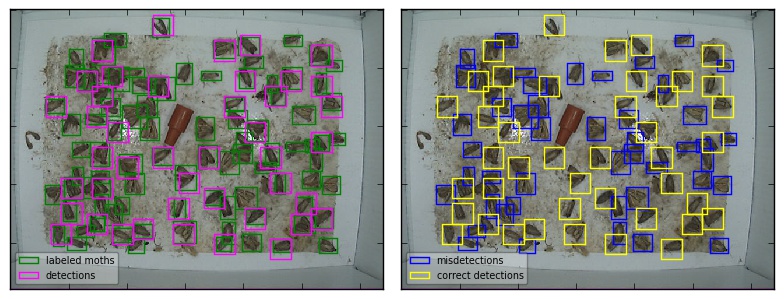}

\includegraphics[width=0.45\textwidth]{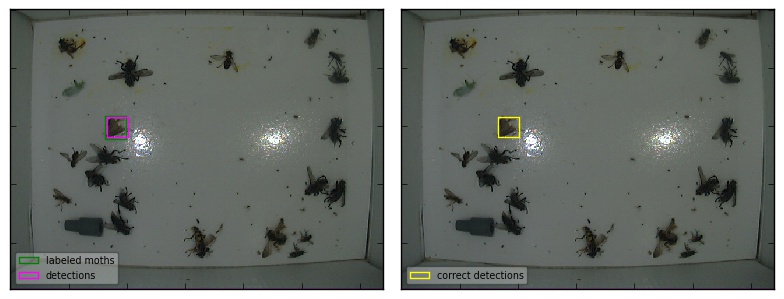}
~
\includegraphics[width=0.45\textwidth]{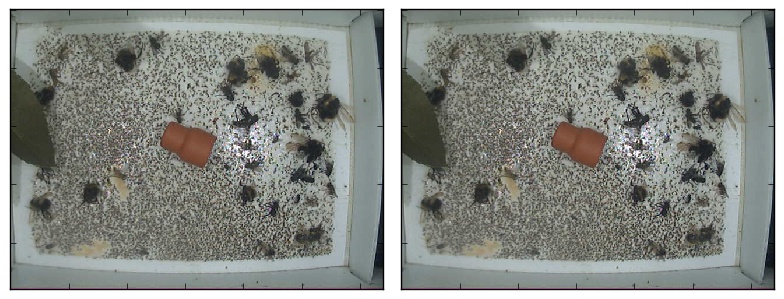}

\includegraphics[width=0.45\textwidth]{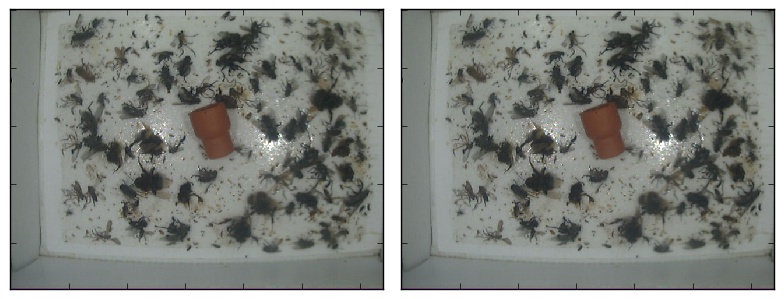}
~
\includegraphics[width=0.45\textwidth]{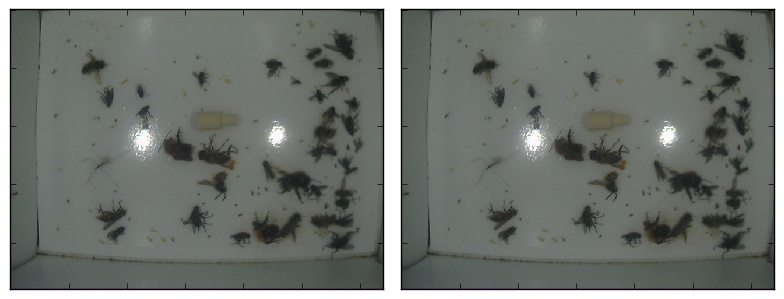}

\caption{More examples of ConvNet with input size $35\times35$. Best viewed in colour.}
\label{fig:more_det_examples_imgthresh}
\end{figure}

Figure \ref{fig:det_examples} shows an example of our detector in operation.
In both panels, the image on the left shows manually annotated bounding boxes in green
and the proposals of our detector in magenta.
The image on the right shows the results of matching annotations with proposals.
Misdetections, false positives and correct detections are
shown in blue, red, and yellow boxes respectively.
In Figure \ref{fig:det_examples_objthresh} and
\ref{fig:det_examples_imgthresh},
the thresholds are set to maximize the F2-score at the object level
and the image level respectively.
Figures \ref{fig:more_det_examples_objthresh} and
\ref{fig:more_det_examples_imgthresh} show more
examples of detection results on full-sized images.

\subsection{Quantitative}
\label{sec:results_quantitative}

\begin{table}[t]
\centering
\begin{tabular}{cccccc}

\hline
\multirow{2}{*}{Method} & \multirow{2}{*}{Patch size} & \multicolumn{2}{c}{Object level} & \multicolumn{2}{c}{Image Level} \\
\cline{3-6}
   & & prec-rec AUC & log-avg miss rate & prec-rec AUC & sens-spec AUC \\
\hline
\multirow{5}{*}{ConvNet}& 21$\times$21   & \textbf{0.931} & \textbf{0.099} & 0.972 & 0.888\\
& 28$\times$28  & 0.879 & 0.135 & 0.987 & 0.913\\
& 35$\times$35  &  0.824 & 0.200 & \textbf{0.993} & \textbf{0.932}\\
& 42$\times$42   &  0.713 & 0.315 & 0.989 & 0.925 \\
& 49$\times$49   &  0.612 & 0.404 & 0.988 & 0.922 \\
\hline
\multirow{5}{*}{LogReg}& 21$\times$21   & 0.555 & 0.756 & 0.782 & 0.682 \\
& 28$\times$28  & 0.621 & 0.586 & 0.765 & 0.665\\
& 35$\times$35  &  \textbf{0.658} & \textbf{0.491} & 0.815 & 0.713\\
& 42$\times$42   &  0.582 & 0.496 & \textbf{0.852} & \textbf{0.755}\\
& 49$\times$49   &  0.499 & 0.545 & 0.808 & 0.713\\

\hline
\end{tabular}
\caption{Performance of convolutional neural networks compared to logistic regression with different input sizes. ​Bold values represent the best performance per-method and per-metric.}
\label{tab:quant_methods}
\end{table}

\begin{figure}[t]
\centering
\begin{subfigure}[b]{0.33\textwidth}
\includegraphics[width=\textwidth]{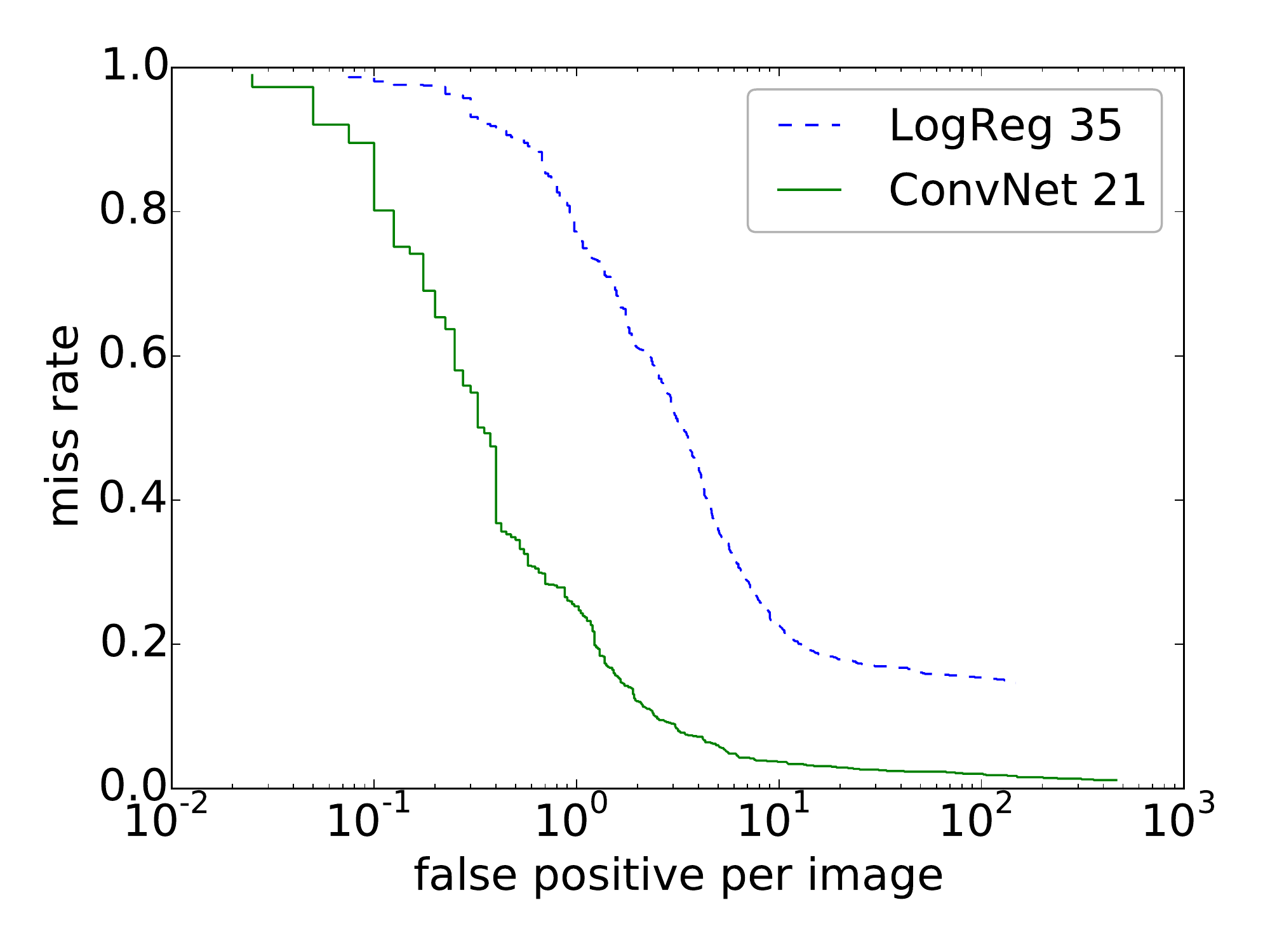}
\caption{miss rate-FPPI at object level}
\end{subfigure}%
\begin{subfigure}[b]{0.33\textwidth}
\includegraphics[width=\textwidth]{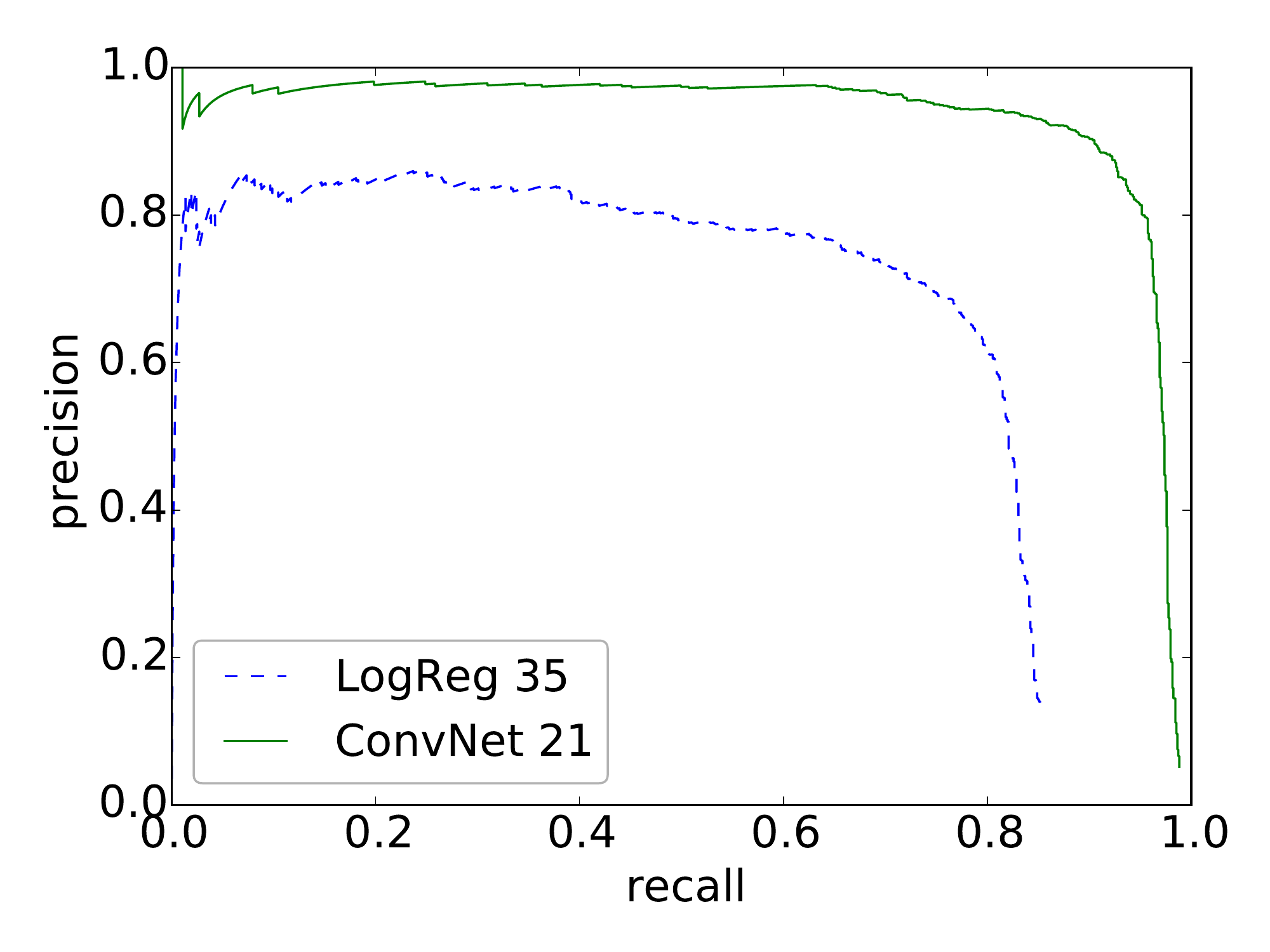}
\caption{precision-recall at object level}
\end{subfigure}%
\begin{subfigure}[b]{0.33\textwidth}
\includegraphics[width=\textwidth]{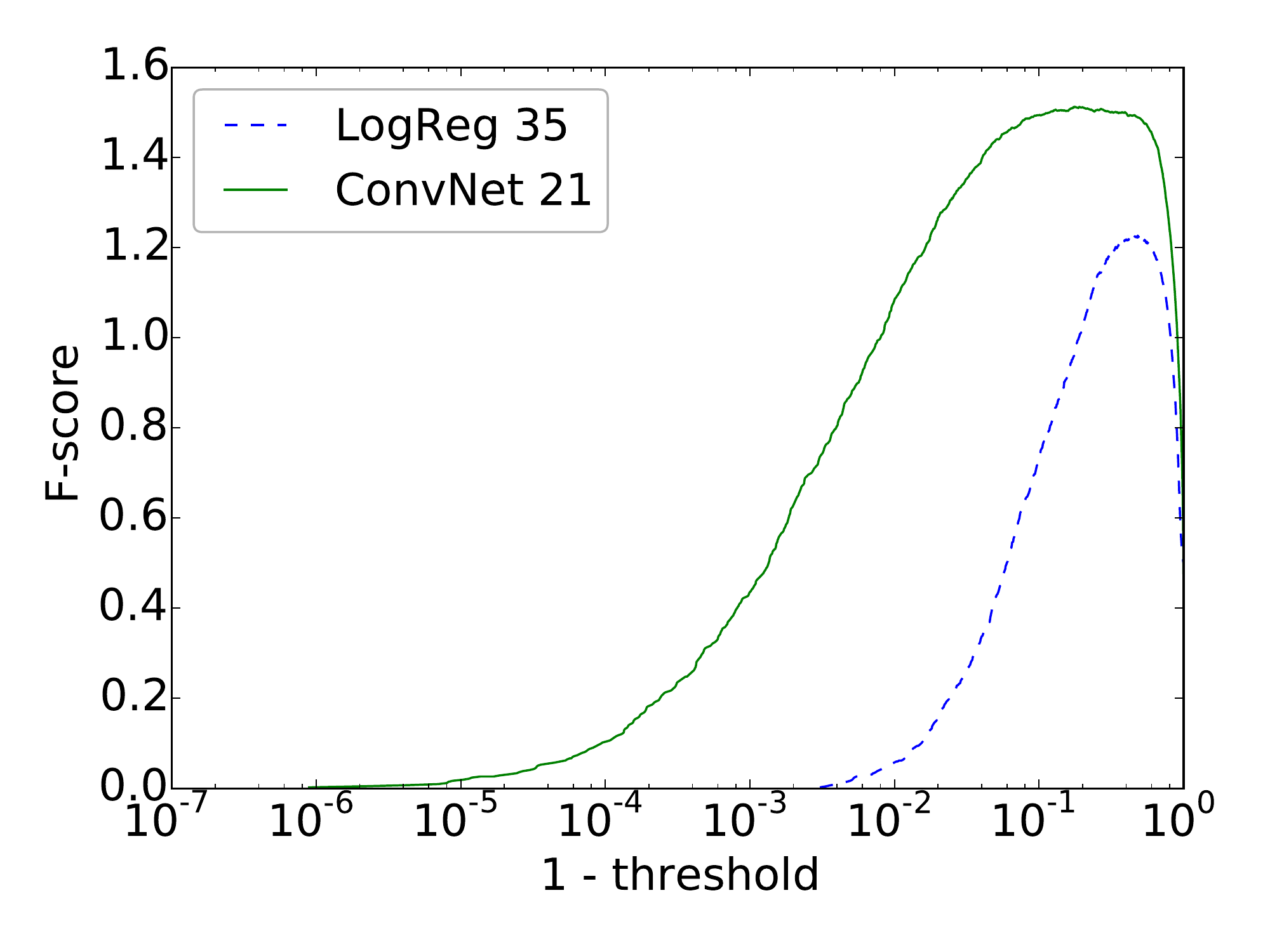}
\caption{$F_2$ score-threshold at object level}
\end{subfigure}%

\begin{subfigure}[b]{0.33\textwidth}
\includegraphics[width=\textwidth]{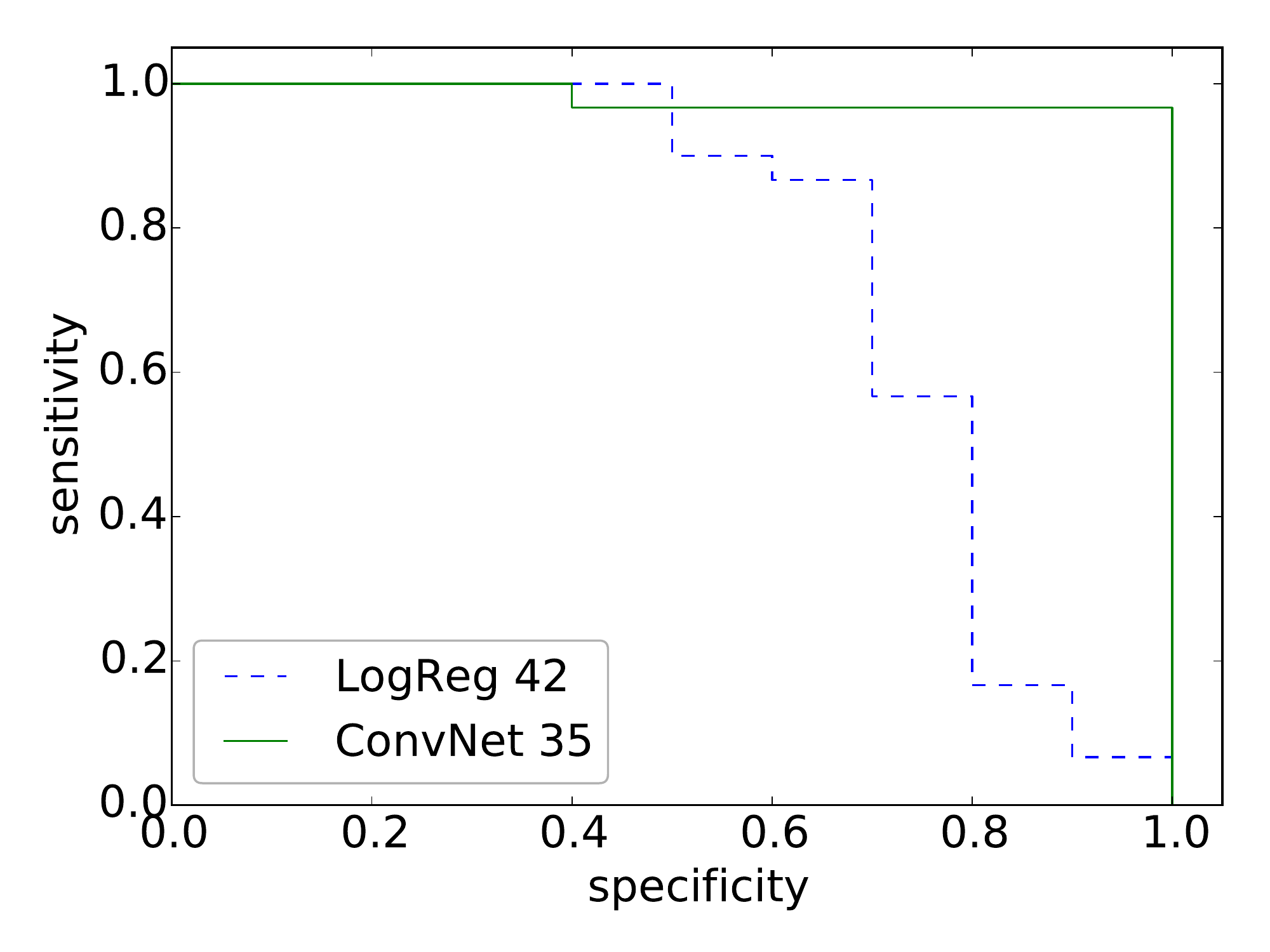}
\caption{sensitivity-specificity at image level}
\end{subfigure}%
\begin{subfigure}[b]{0.33\textwidth}
\includegraphics[width=\textwidth]{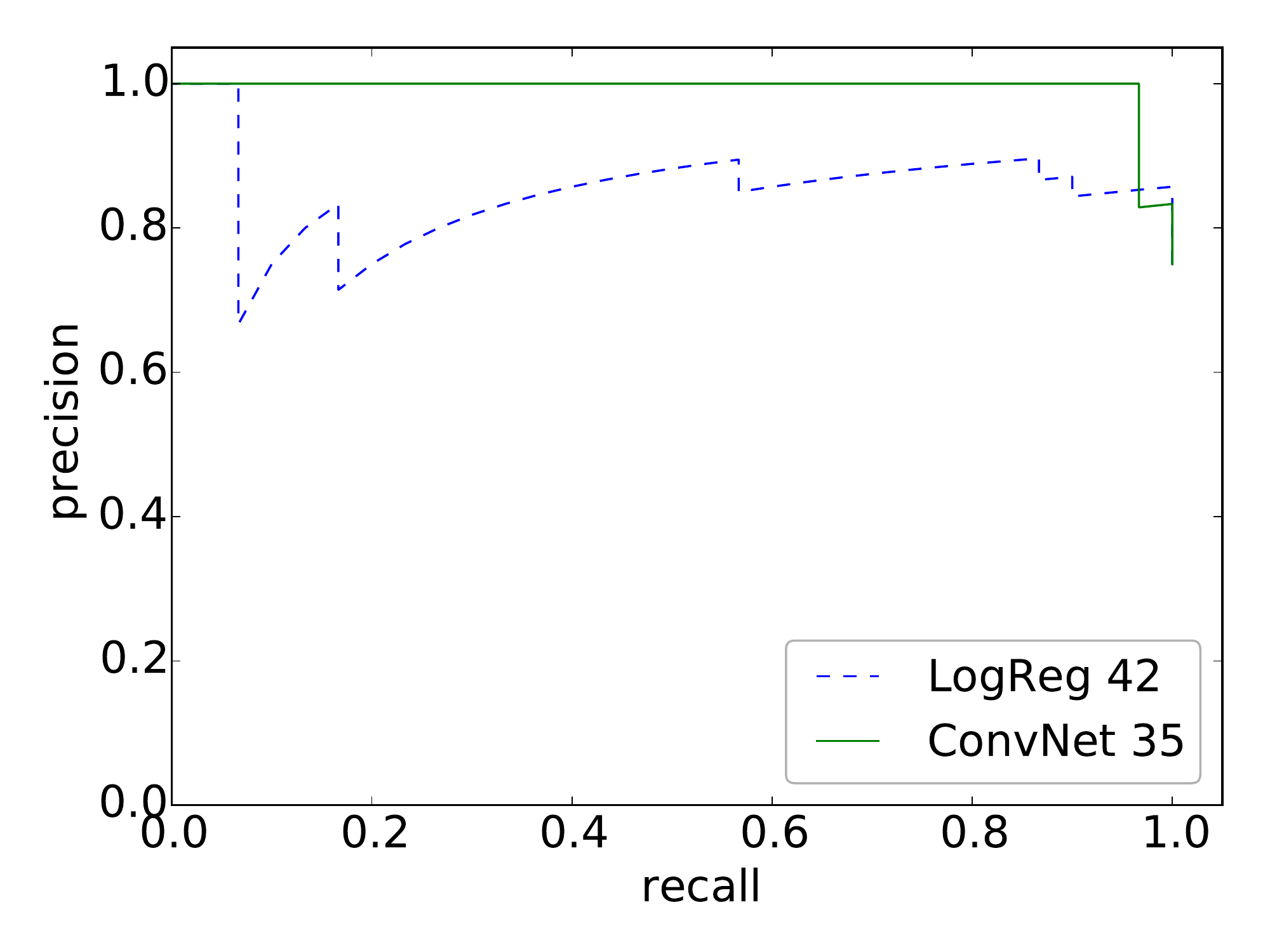}
\caption{precision-recall at image level}
\end{subfigure}%
\begin{subfigure}[b]{0.33\textwidth}
\includegraphics[width=\textwidth]{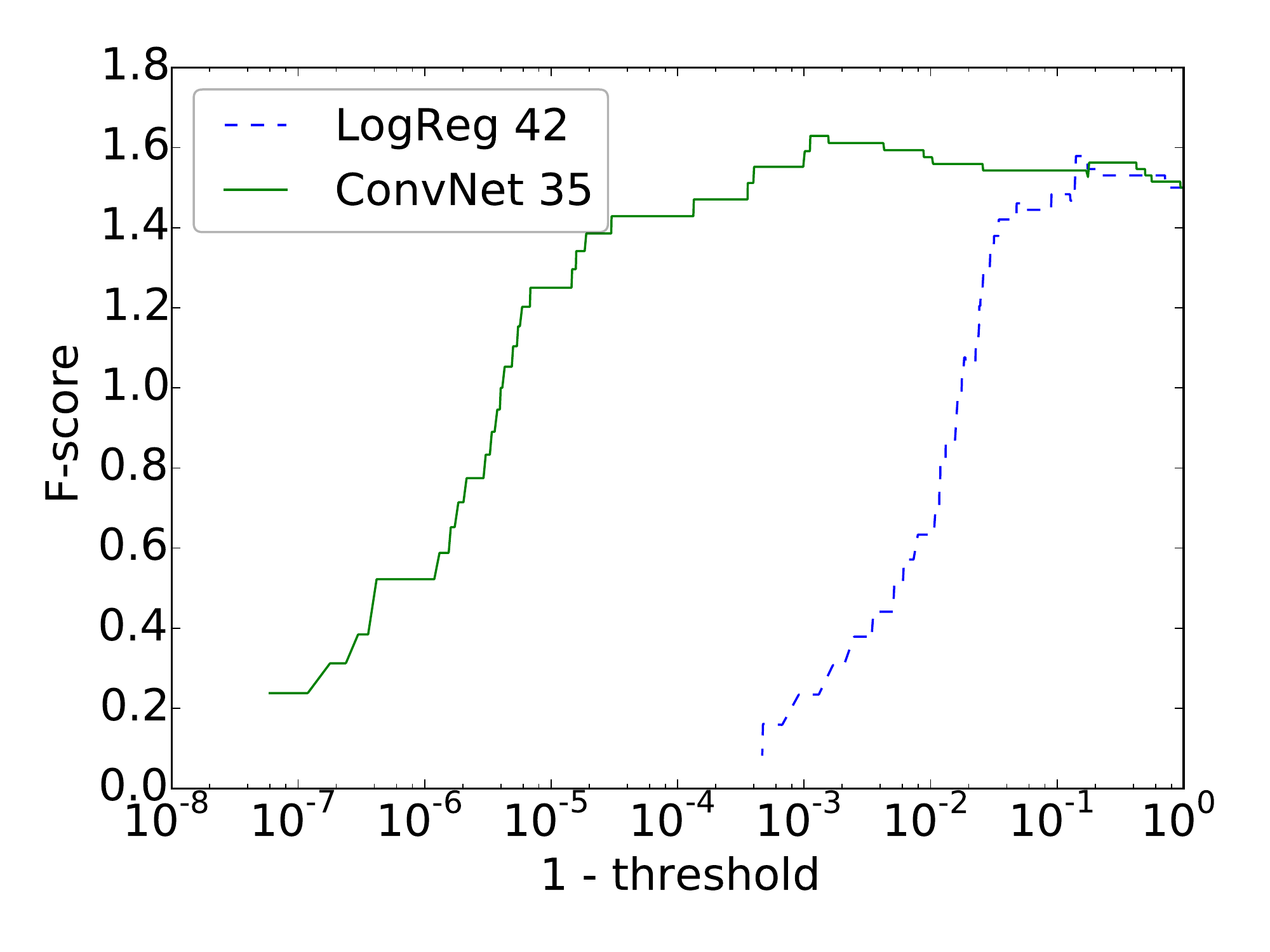}
\caption{$F_2$ score-threshold at image level}
\end{subfigure}%
\caption{Comparison between the best performing ConvNets and logistic
  regression models at both the object and image level.
}
\label{fig:quant}
\end{figure}

\begin{table}[t]
\centering
\begin{tabular}{ccccc}

\hline
\multirow{2}{*}{Method} & \multicolumn{2}{c}{Object Level} & \multicolumn{2}{c}{Image Level} \\
\cline{2-5}
   & prec-rec AUC & log-avg miss rate & prec-rec AUC & sens-spec AUC \\
\hline
ConvNet + Trans\&Rot Aug   & \textbf{0.931} & \textbf{0.099} & 0.972 & 0.888\\
ConvNet + Rot Aug  & 0.916 & 0.140 & \textbf{0.981} & \textbf{0.900}\\
ConvNet + Trans Aug  & 0.879 & 0.204 & 0.979 & 0.892\\
ConvNet + No Aug   & 0.825 & 0.334 & 0.920 & 0.797\\

\hline
\end{tabular}
\caption{Effectiveness of data augmentation.}
\label{tab:quant_aug}
\end{table}

\begin{table}[t]
\centering
\begin{tabular}{ccccc}

\hline
Remaining & \multicolumn{2}{c}{Object Level} & \multicolumn{2}{c}{Image Level} \\
\cline{2-5}
percentage & prec-rec AUC & log-avg miss rate & prec-rec AUC & sens-spec AUC \\
\hline
100\%  & \textbf{0.931} & \textbf{0.099} & 0.972 & 0.888\\
80\%  & 0.923 & 0.121 & \textbf{0.984} & \textbf{0.905}\\
60\%  & 0.906 & 0.155 & 0.979 & 0.883\\
40\% & 0.886 & 0.184 & 0.969 & 0.882\\
20\% & 0.895 & 0.233 & 0.941 & 0.835\\
\hline
\end{tabular}
\caption{Influence of missing training data.}
\label{tab:missing_data}
\end{table}

We chose logistic regression as a baseline\footnote{We also tried a
  popular vision pipeline of local feature descriptors (SIFT
  \cite{lowe2004distinctive}), followed by bag of visual words and a
  support vector machine classifier. This approach did not give
  reasonable detection performance and is thus not reported.}  for
comparison to ConvNets. We tested both logistic regression and
ConvNets at five different input sizes: 21$\times$21, 28$\times$28,
35$\times$35, 42$\times$42, and 49$\times$49.  The results are shown
in Table \ref{tab:quant_methods}.  The ConvNet with input size
21$\times$21 achieved the best performance at the object level and the
ConvNet with input size 35$\times$35 had the best performance at the
image level.  Accordingly, Figure \ref{fig:quant} shows different
performance curves comparing the best performing ConvNet and logistic
regression, both at the object and image level.  Apparent from Figure
\ref{fig:quant} (d) and (e), the ConvNet achieved nearly perfect
results at the image level.  For the precision-recall curve, one
usually expects precision to decrease as recall
increases. Here, in Figure \ref{fig:quant} (b) and (e), precision
sometimes increases as the recall increases.  This is because when the
threshold is decreasing, it is possible that the newly included
detections are all true detections, which results in an increase
in both precision and recall.

To understand the effect of data augmentation (Section
\ref{sec:data_aug}) on detector performance, we performed experiments
on the ConvNet with input size 21$\times$21 by either using (1) both
rotational and translational augmentation; (2) only rotational
augmentation; (3) only translational augmentation; and (4) no
augmentation.  The results are shown in Table \ref{tab:quant_aug}.  We
observed that both translational and rotational augmentation improved
the performance compared to no augmentation at all. Using both
translational and rotational augmentation improved performance at the
object level but not at the image level where a single type of
augmentation was sufficient.

We also evaluated the performance of the proposed method under limited
training data, as shown in Table \ref{tab:missing_data}.
We can see that the algorithm maintains a reasonable performance
even when 80\% of the training data are removed. This also indicates the
effectiveness of the data augmentation strategy.

Some of the moths in the dataset are occluded by other moths, which
increased the difficulty of detection. If we completely remove
occlusion from the dataset, by removing both occluded ground truths
and detections with more than 50\% overlapping with any of the
occluded ground truths during evaluation, we achieve a slight
performance improvement at the object level. Here, the
precision-recall AUC increases from 0.931 to 0.934, and the
log-average miss rate decreases from 0.099 to 0.0916.  This indicates
that our algorithm will perform even better in well-managed sites, where
trap liners are changed often, resulting in less occlusions.

\subsection{Individual detection results}
\label{sec:patch_variability}
\begin{figure}[t]
\centering

\begin{subfigure}[b]{1\textwidth}
\includegraphics[width=0.16\textwidth]{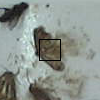}
\includegraphics[width=0.16\textwidth]{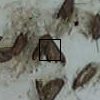}
\includegraphics[width=0.16\textwidth]{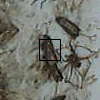}
\includegraphics[width=0.16\textwidth]{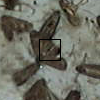}
\includegraphics[width=0.16\textwidth]{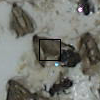}
\includegraphics[width=0.16\textwidth]{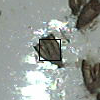}

\includegraphics[width=0.16\textwidth]{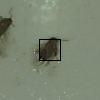}
\includegraphics[width=0.16\textwidth]{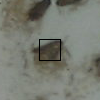}
\includegraphics[width=0.16\textwidth]{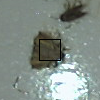}
\includegraphics[width=0.16\textwidth]{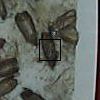}
\includegraphics[width=0.16\textwidth]{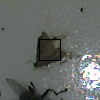}
\includegraphics[width=0.16\textwidth]{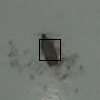}
\caption{Examples of correct detections}
\label{fig:ex_tps}
\end{subfigure}

\vspace{0.5cm}
\begin{subfigure}[b]{1\textwidth}
\includegraphics[width=0.16\textwidth]{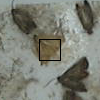}
\includegraphics[width=0.16\textwidth]{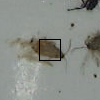}
\includegraphics[width=0.16\textwidth]{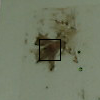}
\includegraphics[width=0.16\textwidth]{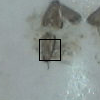}
\includegraphics[width=0.16\textwidth]{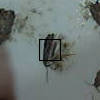}
\includegraphics[width=0.16\textwidth]{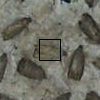}

\includegraphics[width=0.16\textwidth]{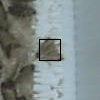}
\includegraphics[width=0.16\textwidth]{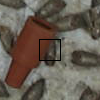}
\includegraphics[width=0.16\textwidth]{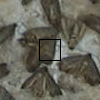}
\includegraphics[width=0.16\textwidth]{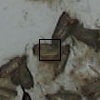}
\includegraphics[width=0.16\textwidth]{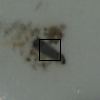}
\includegraphics[width=0.16\textwidth]{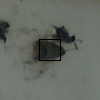}
\caption{Examples of misdetections}
\label{fig:ex_fns}
\end{subfigure}

\vspace{0.5cm}
\begin{subfigure}[b]{1\textwidth}
\includegraphics[width=0.16\textwidth]{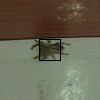}
\includegraphics[width=0.16\textwidth]{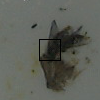}
\includegraphics[width=0.16\textwidth]{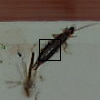}
\includegraphics[width=0.16\textwidth]{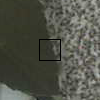}
\includegraphics[width=0.16\textwidth]{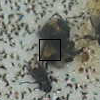}
\includegraphics[width=0.16\textwidth]{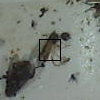}

\includegraphics[width=0.16\textwidth]{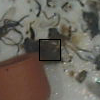}
\includegraphics[width=0.16\textwidth]{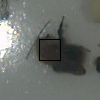}
\includegraphics[width=0.16\textwidth]{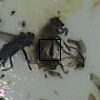}
\includegraphics[width=0.16\textwidth]{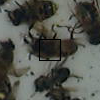}
\includegraphics[width=0.16\textwidth]{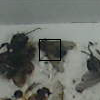}
\includegraphics[width=0.16\textwidth]{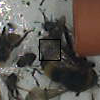}
\caption{Examples of false positives}
\label{fig:ex_fps}
\end{subfigure}

\caption{Image patches (input + context) of correct detections, misdetections
and false positives. Input to the classifier is shown as a black square. 
Best viewed in colour.}
\label{fig:variability}
\end{figure}

Figure \ref{fig:variability} shows various image patches
with different detection outcomes,
including Figure \ref{fig:ex_tps} which shows correct detections,
\ref{fig:ex_fns} which shows misdetections and \ref{fig:ex_fps} which shows false positives.
These patches are all at 100$\times$100 resolution.
They are extracted based on the detection results using an input size
of 21$\times$21.
We can see that moth images show a high degree of variability due to multiple factors,
including different wing poses,
occlusion by other objects,
different decay conditions,
different illumination conditions,
different background textures,
and different blurring conditions.
Some of these moths are successfully detected under these distorting factors
and some are ignored by the detector.
From Figure \ref{fig:ex_fps}, we can also see that inside the 21$\times$21
window considered by the classifier, some of the false positives are, to some extent, visually similar to
the 21$\times$21 image patches that actually are moths. Although for a human reader,
it seems easier to distinguish moth vs.~non-moth by looking at the entire
100 $\times$ 100 patch (i.e.~by considering context).
This suggests that incorporating information from the peripheral region could help
improve detection performance.

\section{Discussion}
\label{sec:discussion}

Compared to the majority of previous work,
the proposed method relies more on data,
and less on human knowledge. No knowledge about codling moths was
considered in the design of our approach. The network learned to
identify codling moths based only on positive and negative training
examples.  This characteristic makes it easier for the system to adapt
to new pest species and new environments, without much manual effort,
as long as relevant data is provided.

Errors caused by different factors were analysed in
Section \ref{sec:patch_variability}.
Many of them are related to time.
The same moth could have different wing poses, levels of occlusion,
illumination and decay conditions over time.
Visual texture can also be related to time. For example, decaying insects could
make the originally white trap liner become dirty, and reduce the contrast between
moths and background.
Errors caused by time-related factors could be largely avoided in real production
systems, if temporal image sequences are provided with reasonable frequency.
This leads to one possible future research direction, that is, to
reason over image sequences while detecting moths on a single image, exploiting
temporal correspondence.
Errors caused by blurry images could potentially be solved by adding
deblurring filters in the preprocessing pipeline.
Non-moth objects contribute to a certain amount of false positives.
One way to address this problem is to train detectors for common non-moth objects
and combine it with the moth detector. Common objects here include pheromone
lures, flies and leaves. This would also require a dataset with richer labelled
information.

As a preliminary attempt on automatic pest detection from trap images,
the methods introduced in this paper have many possible future
extensions besides those have been mentioned based on error analysis.
Deeper convolutional networks \cite{szegedy2014going}
could be applied to provide more accurate image patch classification.
Detecting and classifying multiple types of insects would be a natural
extension, which is closely related to the fine-grained image
classification problem \cite{wang2014learning,razavian2014cnn}.  The
location information of detections could potentially be refined by
proposing rectangular bounding boxes, polygons, or parameterized
curves representing insect shapes.

\section{Conclusions}
\label{sec:conclusions}

This paper describes an automatic method for monitoring pests from
trap images.  We propose a sliding window-based detection pipeline,
where a convolutional neural network is applied to image patches at
different locations to determine the probability of containing a
specific pest type. Image patches are then filtered by non-maximum
suppression and thresholding, according to their locations and
associated confidences, to produce the final detections.
Qualitative and quantitative experiments demonstrate the effectiveness
of the proposed method on a codling moth dataset. We also analysed detection
errors, with corresponding influences to real production systems and
potential future directions for improvements.


\section*{Acknowledgements}
This work was funded by the Natural Sciences and
Engineering Research Council (NSERC) EGP 453816-13, EGP
453816-14, and an industry partner whose name was withheld by
request. We would also like to thank Dr.~Rebecca Hallett, Jordan
Hazell and the industry partner for assistance with data collection.




\bibliographystyle{elsarticle-num}
\bibliography{semios}







\end{document}